\documentclass{article}

\usepackage[main, final]{neurips_2026}

\usepackage[utf8]{inputenc} 
\usepackage[T1]{fontenc}    
\usepackage{hyperref}       
\usepackage{url}            
\usepackage{booktabs}       
\usepackage{amsfonts}       
\usepackage{nicefrac}       
\usepackage{microtype}      
\usepackage{xcolor}         

\usepackage{enumitem} 
\usepackage{graphicx}
\usepackage{subcaption}
\usepackage{siunitx}
\usepackage{multirow}

\usepackage{amsmath}
\usepackage{amssymb}
\usepackage{mathtools}
\usepackage{amsthm}
\usepackage{bbm}
\usepackage[capitalize,noabbrev]{cleveref}

\theoremstyle{plain}
\newtheorem{theorem}{Theorem}[section]

\theoremstyle{definition}
\newtheorem{definition}[theorem]{Definition}

\theoremstyle{remark}

\definecolor{custombeige}{HTML}{F4C7A8}
\definecolor{customtail}{HTML}{4fb9af}

\usepackage{bm}
\usepackage{braket}
\usepackage[T1]{fontenc}

\DeclareMathOperator{\kl}{\mathrm{kl}}

\usepackage{colortbl}

\definecolor{blue1}{RGB}{204,229,255}
\definecolor{blue2}{RGB}{0,76,153}

\usepackage{pifont}
\newcommand{\cmark}{\ding{51}}%
\newcommand{\xmark}{\ding{55}}%

\usepackage{tcolorbox}
\tcbuselibrary{skins, breakable}
\tcbset{
    boxrule=0pt,
    colframe=gray!90,
    colback=gray!20,
    coltitle=white,
    enhanced,
    boxsep=5pt,
    arc=2mm}

\usepackage{colortbl}
\usepackage{pifont}

\definecolor{btight}{HTML}{1B7837}
\definecolor{bgood}{HTML}{7FBF7B}
\definecolor{bok}{HTML}{F0E442}
\definecolor{bloose}{HTML}{F4A582}
\definecolor{bvac}{HTML}{CA0020}

\newcommand{\btight}[1]{\cellcolor{btight!25}\textbf{#1}}
\newcommand{\bgood}[1]{\cellcolor{bgood!35}{#1}}
\newcommand{\bok}[1]{\cellcolor{bok!40}{#1}}
\newcommand{\bloose}[1]{\cellcolor{bloose!40}{#1}}
\newcommand{\bvac}[1]{\cellcolor{bvac!30}{#1}}

\title{Escaping Parameter Space: Tight Generalization Bounds via Representation Quality}

\author{%
  Niclas G\"oring$^{1}$\thanks{Corresponding author: \texttt{niclas.goring@physics.ox.ac.uk}}\quad
  Shuofeng Zhang$^{1}$\quad
  Branton DeMoss$^{2}$\quad
  Ard A.~Louis$^{1}$\\[4pt]
  $^{1}$Rudolf Peierls Centre for Theoretical Physics, University of Oxford, Oxford, UK\\
  $^{2}$Mathematical Institute, University of Oxford, Oxford, UK
}

\begin{document}

\maketitle

\begin{abstract}
We derive a tight generalization bound for deterministic neural We derive a tight generalization bound for deterministic neural networks based on their last hidden layer representations. By evaluating representation quality with a parameter-free $k$-nearest-neighbor ($k$-NN) classifier, our certificate avoids traditional parameter-space complexity measures, KL divergences to a posterior, weight norms, and compression. The bound is tight: trained from scratch, it reaches an 8.0 percentage point gap to test error on CIFAR-10 (ResNet-18) and yields the first non-vacuous certificates we are aware of on CIFAR-100; with frozen DINOv2 representations, it certifies at 1.4\% on CIFAR-10 and 23.1\% on 1,000-class ImageNet. It decomposes additively into three empirical, independently measurable quantities: local class geometry ($k$-NN decoder error), decoder-network agreement, and training stability. This allows the bound to act as a diagnostic tool: we demonstrate that actively destroying local class geometry eliminates generalization while leaving dataset memorization intact, identifying class geometry as a causally relevant mechanism for generalization. Finally, we show that learned representations are significantly less sensitive to weight noise than linear readouts, suggesting  why parameter-space certificates often struggle to remain tight.

\end{abstract}

\section{Introduction}
\label{sec:introduction}

Almost every existing non-vacuous generalization bound for deep networks pays a price
for their treatment of  parameter space. PAC-Bayes certificates add noise to the weights and pay
through KL divergence~\cite{dziugaite2017computingnonvacuousgeneralizationbounds,pérezortiz2021tighterriskcertificatesneural,lotfi2022pacbayescompressionboundstight}.
Compression bounds quantize or prune and pay through description
length~\cite{zhou2019nonvacuousgeneralizationboundsimagenet,lotfi2022pacbayescompressionboundstight}.
Recent work has
produced non-vacuous certificates for deterministic networks by reasoning about
partitions~\cite{than2025nonvacuousgeneralizationboundsdeep} or distribution and
loss complexity. 
Each line of work has produced progressively tighter
results, but all share a common design: they reason about parameters, or
parameter-derived quantities, to certify predictions.

\begin{figure}[htb!]
  \centering
  \includegraphics[width=\textwidth]{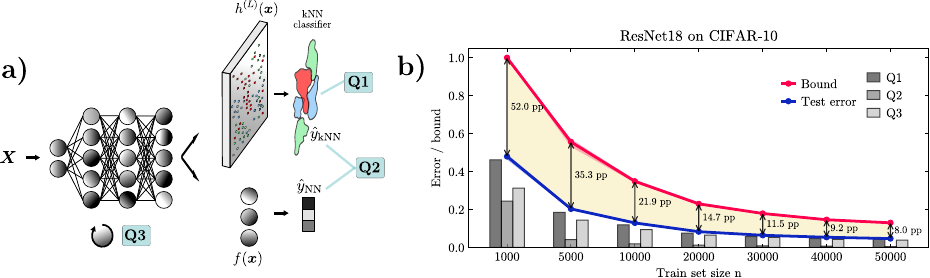}
  \caption{
  Any learned feature map $\varphi\colon \bm X\to\mathcal{Z}$ raises three questions:
  \textbf{Q1}~Can a simple classifier in $\mathcal{Z}$ predict labels?
  \textbf{Q2}~Does it agree with the network?
  \textbf{Q3}~Is training stable under fold removal?
  \textbf{a)}~A $k$-NN classifier on the last-hidden-layer representation $h^{(L)}(\boldsymbol{x})$ yields $\hat{y}_{\mathrm{kNN}}$~(\textbf{Q1}), compared against $\hat{y}_{\mathrm{NN}}$~(\textbf{Q2}); a fold-trained model tests stability~(\textbf{Q3}).
  \textbf{b)}~ResNet-18 on CIFAR-10: the bound converges to within $8.0$\,pp of test error at $n{=}50{,}000$, the tightest known certificate from a deterministic network without modified training. Stacked bars show \textbf{Q1} dominates: representation quality is the binding constraint.}
  \label{fig:overview}
\end{figure}

We take a different route. Instead of certifying the parameterized network through its weights, we certify it through its learned representation. Concretely, we produce two bounds: in the transfer setting, where the feature map is fixed independently of the target data, we certify a
$k$-nearest-neighbor ($k$-NN) classifier built on the last-hidden-layer representations $h^{(L)}(\bm{x})$; in the
cross-fit setting, where the network is trained on the target task, we certify the deployed deterministic network itself, using the
$k$-NN decoder. The proof uses only cross-fitting and the KL concentration inequality for i.i.d.\ Bernoulli variables. The representation is of course produced by the parameters, but no parameter-space complexity measure, no KL to a posterior, weight norm, covering number, or description length enters the argument. The bound is a function of three empirical quantities on the learned
feature map $\varphi\colon \bm X\to\mathcal{Z}$
(Figure~\ref{fig:overview}a):
\begin{enumerate}[label=\textbf{Q\arabic*},leftmargin=3em,itemsep=0pt]
\item \label{q:decode}
  \emph{Can a simple classifier in $\mathcal{Z}$ predict labels?}
  That is, does the representation have clean local class geometry?
\item \label{q:agree}
  \emph{Does that classifier agree with the network?}
  That is, does the representation capture what the network actually computes?
\item \label{q:stable}
  \emph{Is training stable under fold removal?}
  That is, do models trained on $n$ versus $n-n_j$ points make essentially the
  same predictions?
\end{enumerate}

When all three answers are yes, the deterministic deployed network is certified
without ever reasoning about its parameters. The bound decomposes additively
into one term per question, turning the certificate into a diagnostic of
representation quality. Our \textbf{contributions} are:
\begin{enumerate}[leftmargin=2em,itemsep=2pt,topsep=2pt]

\item \textbf{A representation-level certificate for deterministic
  networks} (\cref{cor:full}). Three empirical terms, decoder error
  (\ref{q:decode}), decoder--network disagreement (\ref{q:agree}), and
  training stability (\ref{q:stable}) bound the test error of the
  deployed network, without any parameter-space KL term, weight norm, covering number, or
  compression in the proof.

\item \textbf{Tight bounds across scales.} Trained from scratch: gaps
  to test error as small as $8.0$\,pp obtain on CIFAR-10, and the first
  non-vacuous certificates we are aware of on CIFAR-100 for
  unmodified deterministic networks. Frozen pretrained representations achieve
  $1.4\%$ on CIFAR-10 and $23.1\%$ on 1{,}000-class ImageNet (DINOv2),
  the first non-vacuous $k$-NN decoder certificates we are aware of at
  ImageNet scale. See \cref{tab:transfer,tab:comparison_merged}.

\item \textbf{Implications of the decomposition.} The three terms can be
  measured independently, which localizes where generalization succeeds
  or fails. Double descent lives in the linear readout (\ref{q:agree}), not the
  representation (\ref{q:decode}); see \cref{sec:overparam}. An intervention that destroys
  local class geometry eliminates generalization while training accuracy
  stays at $100\%$,
  interventional evidence that \ref{q:decode} is causal, not a
  correlate of generalization (\cref{sec:causal}).

\end{enumerate}

\subsection{Related work}
\label{sec_relatedwork}

Most non-vacuous certificates for deep networks reason through parameter-space
quantities, whether via PAC-Bayes perturbation of the
weights~\citep{mcallester1998some,dziugaite2017computingnonvacuousgeneralizationbounds,neyshabur2017pac,pérezortiz2021tighterriskcertificatesneural}
or compression and
quantization~\citep{hinton1993keeping,arora2018stronger,zhou2019nonvacuousgeneralizationboundsimagenet,lotfi2022pacbayescompressionboundstight}.
Our bound takes a different route: no parameter-space complexity measure enters the certificate, and its tightness depends on representation quality rather than parameter count (\cref{sec:overparam}). Function-space PAC-Bayes analyses also move beyond
parameter space~\citep{valle2020generalization,zhang2025position};
our approach is not the first to do so, but requires only held-out error,
agreement, and stability. Two recent methods are related:
\citep{than2025nonvacuousgeneralizationboundsdeep} certify deterministic
networks via an a~priori data partition;
\citep{bazinet2026bounddisagreegeneralizationbounds} certify via a
separately trained surrogate and empirical disagreement. Our surrogate
is neither trained nor chosen: it is a parameter-free $k$-NN decoder
induced by the representation itself, and the bridge to the deployed
network reduces to training
stability~\citep{bousquet2002stability,hardt2016train}. Our proof uses cross-fit held-out evaluation with KL
concentration~\citep{langford2005tutorial,blum1999beating}; our contribution is
not the technique but the object being evaluated.
$k$-NN probing of representations is standard in self-supervised
learning~\citep{wu2018unsupervised,caron2021emerging,oquab2023dinov2}; our
contribution is that probe performance, combined with agreement and stability,
becomes a formal certificate for the deployed network. We compare numerical values of our certificates in \cref{tab:transfer,tab:comparison_merged} and discuss related work in more depth in Appendix~\ref{appx_relwork}.

\section{Theory: A Representation-Level Bound}
\label{sec:certificate}
Any learned feature map $\bm\varphi\colon\bm X\to\mathcal{Z}$ turns a
dataset into a point cloud in some metric space $(\mathcal{Z},d)$. In practice,
$\bm\varphi$ is the network's last hidden layer representation $h^{(L)}(\bm{x})$, the vector fed into the final
linear classifier, and $d$ is $\ell_2$ distance; extraction requires a single
forward pass with no modification to the network (details in
Appendix~\ref{appx:repr_details}). The theorems apply verbatim to any
deterministic feature map (e.g. another layer of the network) and metric $d$. We instantiate the three questions from the introduction by taking the ``simple
classifier'' to be a $k$-NN rule in $(\mathcal{Z},d)$. It
requires no training, has no parameters of its own, and makes all three
quantities directly measurable. We emphasize what this does and does not mean: the representation $\bm\varphi$ is of course produced by the trained parameters, but the bound uses no parameter-space complexity measure. The proofs use a single structural property:
conditioned on the complement sample, the held-out points are i.i.d.\ and the
decoder is deterministic, so standard concentration applies.

\subsection{Technical setup}
\label{subsec:setup}

Let $P$ be a distribution on $\bm X \times [C]$, where
$[C] \coloneqq \{1,\dots,C\}$. Draw an i.i.d.\ sample
$\mathcal{S}_n = \{(\bm x_i, y_i)\}_{i=1}^{n} \sim P^n$. Let
$\mathcal{A}$ be a (possibly randomized) training algorithm that maps any
labeled dataset~$\mathcal{T}$ to network parameters
$\mathcal{A}(\mathcal{T})$, producing a classifier
$\widehat{f}_{\mathcal{A}(\mathcal{T})}$ and a feature map
$\bm\varphi_{\mathcal{A}(\mathcal{T})} \colon \bm X \to \mathcal{Z}$
into a metric space $(\mathcal{Z}, d)$. All internal randomness of
$\mathcal{A}$ (initialization, data ordering, augmentation) is fixed in advance;
every statement below is conditional on this randomness. Write
$\bm\theta^{(\mathrm{full})} = \mathcal{A}(\mathcal{S}_n)$ for the full-data
parameters, $\widehat{f}^{(\mathrm{full})}$ for the deployed classifier, and
$\bm\varphi^{(\mathrm{full})} = \bm\varphi_{\bm\theta^{(\mathrm{full})}}$ for
its feature map. This is the object we wish to certify.

\begin{definition}[Cross-fit $k$-NN decoder]
\label{def:crossfit}
Partition $[n]$ into $J$ disjoint folds $F_1,\dots,F_J$ with $|F_j| = n_j$ and
$\sum_j n_j = n$. For each fold~$j$, train on the complement:
\[
  \bm\theta^{(-j)} = \mathcal{A}\!\bigl(\mathcal{S}_{-j}\bigr),
  \qquad
  \mathcal{S}_{-j} = \bigl\{(\bm x_i,y_i) : i \notin F_j\bigr\}.
\]
Write $\bm\varphi^{(-j)}$ for the fold-$j$ feature map,
$\bm z_i^{(-j)} = \bm\varphi^{(-j)}(\bm x_i)$ for the embedded points, and
$\widehat{f}_j = \widehat{f}_{\bm\theta^{(-j)}}$ for the fold-$j$ classifier.
The \emph{cross-fit $k$-NN decoder} $h_j^{(k)}$ maps a query~$\bm x$ to the
majority label among the $k$ nearest neighbors of
$\bm\varphi^{(-j)}(\bm x)$ in the reference set
$\{\bm z_i^{(-j)} : i \notin F_j\}$, with distance~$d$. All tie-breaking rules
are fixed in advance and deterministic.
\end{definition}

Three empirical quantities control the certificate:
\begin{center}\small
\renewcommand{\arraystretch}{1.3}
\begin{tabular}{@{} l c l @{}}
\toprule
 & Definition & Measures \\
\midrule
$\widehat{R}_j^{(k)}$
  & $\frac{1}{n_j}\sum_{i \in F_j}
    \mathbf{1}[h_j^{(k)}(\bm x_i) \neq y_i]$
  & \ref{q:decode}: decoder error \\[4pt]
$\widehat{D}_j^{(k)}$
  & $\frac{1}{n_j}\sum_{i \in F_j}
    \mathbf{1}[h_j^{(k)}(\bm x_i) \neq \widehat{f}_j(\bm x_i)]$
  & \ref{q:agree}: decoder--network gap \\[4pt]
$  \Delta_j^{\mathrm{stab}}$
  & $\frac{1}{n_j}\sum_{i \in F_j}
    \mathbf{1}[\widehat{f}_j(\bm x_i) \neq
    \widehat{f}^{(\mathrm{full})}(\bm x_i)]$
  & \ref{q:stable}: fold--full gap \\
\bottomrule
\end{tabular}
\end{center}
Note that $  \Delta_j^{\mathrm{stab}}$ is independent of~$k$: it compares the two networks'
hard predictions directly.

\paragraph{Notation.}
The binary relative entropy is
$\kl(q \| p)
  = q\log\tfrac{q}{p} + (1{-}q)\log\tfrac{1-q}{1-p}$
for $q,p \in [0,1]$ (extended by continuity at the boundary), and its upper
inverse is
$\kl^{-1}_+(q,c) = \sup\{p \in [q,1] : \kl(q\|p) \le c\}$.

\subsection{Certifying the decoder (\ref{q:decode}) }
\label{subsec:thm_decoder}

The first step certifies the $k$-NN decoder alone, treating it as a
classifier in its own right.  Because the fold points~$F_j$ are
held out from training $\bm\theta^{(-j)}$, the decoder is
conditionally deterministic and the error indicators are conditionally
i.i.d. Hence, a direct application of KL concentration suffices.

\begin{theorem}[Cross-fit decoder certificate]
\label{thm:decoder}
Fix a finite search set $\mathcal{K} \subset \mathbb{N}$
\textup{(}e.g.\ $\mathcal{K} = \{1,3,5\}$\textup{)}.
With probability at least $1 - \delta$ over $\mathcal{S}_n$,
for every $j \in [J]$ and $k \in \mathcal{K}$ simultaneously:
\begin{equation}
\label{eq:decoder}
  \Pr_{(\bm X, Y) \sim P}\!\bigl(h_j^{(k)}(\bm X) \neq Y
  \mid \mathcal{S}_{-j}\bigr)
  \;\le\;
  \kl^{-1}_+\!\left(
    \widehat{R}_j^{(k)},\;\;
    \frac{
      \log\!\bigl(|\mathcal{K}|\,J / \delta\bigr)
      + \log(n_j {+} 1)
    }{n_j}
  \right).
\end{equation}
Since $h_j^{(k)}$ depends on $\mathcal{S}_n$ only through
$\mathcal{S}_{-j}$, conditioning on $\mathcal{S}_{-j}$ is equivalent
to conditioning on $\mathcal{S}_n$.
\end{theorem}

\begin{proof}
See \cref{appx:proof_decoder} for the proof.
\end{proof}


\subsection{Transfer certificate}
\label{subsec:transfer}

When the feature map $\bm\varphi$ is learned without access to the
target data, e.g.\ pretrained on a different dataset, the cross-fit
construction simplifies.  There is no fold/full distinction, so
\ref{q:stable} vanishes, \ref{q:agree} is irrelevant (the $k$-NN
decoder \emph{is} the deployed classifier), and a single
reference/evaluation split replaces the $J$-fold construction. For example, this lets us certify a $k$-NN decoder built from an ImageNet-pretrained ResNet-50 representation on CIFAR-10; see \cref{sec:transfer}.

\begin{tcolorbox}[
  colback=gray!5, colframe=black!80,
  boxsep=2pt, left=6pt, right=6pt, top=4pt, bottom=4pt,
  fonttitle=\normalsize,
  boxrule=0.4pt,
  coltitle=black, colbacktitle=gray!15,
]
\begin{theorem}[Transfer certificate]
\label{prop:transfer}

Let $\bm\varphi$ be fixed independently of the target
sample~$\mathcal{S}_n$ (e.g.\ pretrained on disjoint data).
Split $\mathcal{S}_n \sim P^n$ into $\mathcal{S}_{\mathrm{ref}}$
(size~$m$) and $\mathcal{S}_{\mathrm{eval}}$
(size~$n_e = n{-}m$).  Let $I_{\mathrm{eval}} \subset [n]$ denote
the indices of the evaluation points.
Build $h^{(k)}$ from
$\bm\varphi(\mathcal{S}_{\mathrm{ref}})$.
Define $\widehat{R}^{(k)}$ as in \cref{subsec:setup} but computed
on $\mathcal{S}_{\mathrm{eval}}$.
With probability $\geq 1 - \delta$ over $\mathcal{S}_n$, for every
$k \in \mathcal{K}$ simultaneously:
\begin{equation}
\label{eq:transfer-decoder}
  \Pr_{(\bm X, Y) \sim P}\!\bigl(h^{(k)}(\bm X) \neq Y
    \mid \mathcal{S}_{\mathrm{ref}}\bigr)
  \;\leq\;
  \kl^{-1}_+\!\bigl(\widehat{R}^{(k)},\; c_e\bigr),
\end{equation}
where\;
$c_e = \bigl[\log(|\mathcal{K}|/\delta)
  + \log(n_e{+}1)\bigr]/n_e$.
\end{theorem}
\tcblower
\small
\ref{q:agree}${} = 0$ and \ref{q:stable}${} = 0$ by construction:
the $k$-NN decoder is the classifier and $\bm\varphi$ does not
depend on the target data.
Eq.~\eqref{eq:transfer-decoder} certifies the decoder via
\ref{q:decode} alone.
\end{tcolorbox}

\begin{proof}
See \cref{appx:proof_2} for the proof.
\end{proof}

\subsection{Certifying the full network}
\label{subsec:full_network}

When the network is trained on the target task, \ref{q:agree}
and \ref{q:stable} are both non-trivial.  We handle them in two
steps. We bound the deployed network $\widehat{f}^{(\mathrm{full})}$ without accessing its parameters: its error is at most (a) the error of a surrogate fold network plus (b) the disagreement between fold and full networks on unlabeled data.

\paragraph{Step A: from decoder to fold network.}

\begin{theorem}[Fold-network certificate]
\label{thm:network}
Fix $\mathcal{K} \subset \mathbb{N}$.  With probability at least
$1 - \delta$ over~$\mathcal{S}_n$, for every $j \in [J]$ and
$k \in \mathcal{K}$ simultaneously:
\begin{equation}
\label{eq:network}
  \Pr_{(\bm X, Y) \sim P}\!\bigl(\widehat{f}_j(\bm X) \neq Y
    \mid \mathcal{S}_{-j}\bigr)
  \;\le\;
  \underbrace{
    \kl^{-1}_+\!\bigl(\widehat{R}_j^{(k)},\; c_j\bigr)
  }_{\textup{\ref{q:decode}}}
  \;\;+\;\;
  \underbrace{
    \kl^{-1}_+\!\bigl(\widehat{D}_j^{(k)},\; c_j\bigr)
  }_{\textup{\ref{q:agree}}}
\end{equation}
where\;
$c_j = \bigl[\log(2\,|\mathcal{K}|\,J/\delta)
  + \log(n_j{+}1)\bigr] / n_j$.
Since $\widehat{f}_j$ depends on $\mathcal{S}_n$ only through
$\mathcal{S}_{-j}$, conditioning on $\mathcal{S}_{-j}$ is equivalent
to conditioning on~$\mathcal{S}_n$.
\end{theorem}

\begin{proof}
See \cref{appx:proof_3} for the proof.
\end{proof}

\paragraph{Step B: from fold network to full-data network.}
For any $(\bm x, y)$ and $j \in [J]$,
\begin{equation}
\label{eq:triangle_stab}
  \mathbf{1}\!\bigl[\widehat{f}^{(\mathrm{full})}(\bm x) \neq y\bigr]
  \;\le\;
  \mathbf{1}\!\bigl[\widehat{f}_j(\bm x) \neq y\bigr]
  \;+\;
  \mathbf{1}\!\bigl[\widehat{f}_j(\bm x)
    \neq \widehat{f}^{(\mathrm{full})}(\bm x)\bigr],
\end{equation}
since the full network can only err if the fold network also errs
or the two disagree.
Taking expectations conditional on $\mathcal{S}_n$ introduces the
\emph{stability gap}
\begin{equation}
\label{eq:stab_gap}
  \Delta_j^{\mathrm{stab}}
  \;=\;
  \Pr_{\bm X \sim P_X}\!\bigl(\widehat{f}_j(\bm X)
    \neq \widehat{f}^{(\mathrm{full})}(\bm X)
    \mid \mathcal{S}_n\bigr).
\end{equation}
This is a disagreement probability over $\bm X$ alone; labels play no role, so it can be certified from unlabeled data. Rather than derive an optimizer-specific analytic stability bound that would need to be rewritten for SGD, Adam, RAdam, Lion, Shampoo, or any new optimizer, we certify this term empirically. This keeps the certificate practically usable across training setups.

\begin{theorem}[Direct stability certification]
\label{prop:q3_unlabeled}
Let $\mathcal{U} = \{\bm x_r\}_{r=1}^{m} \sim P_X^{m}$ be
independent of~$\mathcal{S}_n$.\footnote{%
  The standard test inputs of MNIST, CIFAR-10, and CIFAR-100
  satisfy this: no test point participates in training any fold
  or full model.}
Define
$\widehat{\Delta}_j^{\,\mathcal{U}}
  = \frac{1}{m}\sum_{r=1}^{m}
  \mathbf{1}[\widehat{f}_j(\bm x_r) \neq
    \widehat{f}^{(\mathrm{full})}(\bm x_r)]$.
With probability $\ge 1 - \delta'$ over the joint draw
$(\mathcal{S}_n, \mathcal{U})$, for every $j \in [J]$
simultaneously:
\begin{equation}
\label{eq:q3_cert}
  \Delta_j^{\mathrm{stab}}
  \;\le\;
  \kl^{-1}_+\!\left(
    \widehat{\Delta}_j^{\,\mathcal{U}},\;\;
    \frac{\log(J / \delta') + \log(m{+}1)}{m}
  \right)
  \;=:\;
  B_j^{\mathrm{stab}}.
\end{equation}
\end{theorem}

\begin{proof}
    See \cref{appx:proof_stab} for the proof.
\end{proof}

\noindent
Combining \cref{thm:network},
\eqref{eq:triangle_stab}, and
\cref{prop:q3_unlabeled}:

\begin{tcolorbox}[
  colback=gray!5, colframe=black!80,
  boxsep=2pt, left=6pt, right=6pt, top=4pt, bottom=4pt,
  fonttitle=\normalsize,
  boxrule=0.4pt,
  coltitle=black, colbacktitle=gray!15,
]
\begin{theorem}
    \label{cor:full}%
Let $\mathcal{U} = \{\bm x_r\}_{r=1}^{m} \sim P_X^{m}$ be
independent of~$\mathcal{S}_n$.
With probability ${\ge}\; 1-\delta - \delta'$ over
$(\mathcal{S}_n, \mathcal{U})$, for every
$k\!\in\!\mathcal{K}$ and $j\!\in\![J]$:
\begin{equation}
\label{eq:main}
  \Pr_{(\bm X, Y) \sim P}\!\bigl(
    \widehat{f}^{(\mathrm{full})}(\bm X) \neq Y
    \mid \mathcal{S}_n\bigr)
  \;\;\le\;\;
  \underbrace{
    \kl^{-1}_+\!\bigl(\widehat{R}_j^{(k)},\; c_j\bigr)
  }_{\substack{
    \textup{\ref{q:decode}: local class} \\[-2pt]
    \textup{geometry}
  }}
  \;\;+\;\;
  \underbrace{
    \kl^{-1}_+\!\bigl(\widehat{D}_j^{(k)},\; c_j\bigr)
  }_{\substack{
    \textup{\ref{q:agree}: decoder--} \\[-2pt]
    \textup{network agreement}
  }}
  \;\;+\;\;
  \underbrace{
    \kl^{-1}_+\!\bigl(
      \widehat{\Delta}_j^{\,\mathcal{U}},\; c_m\bigr)
  }_{\substack{
    \textup{\ref{q:stable}: training} \\[-2pt]
    \textup{stability}
  }}
\end{equation}
where\;
$c_j = \bigl[\log(2\,|\mathcal{K}|\,J/\delta)
  + \log(n_j{+}1)\bigr] / n_j$
\;and\;
$c_m = \bigl[\log(J/\delta') + \log(m{+}1)\bigr] / m$.
The tightest certificate is
$\min_{j,k}\bigl[
  \kl^{-1}_+(\widehat{R}_j^{(k)}, c_j)
  + \kl^{-1}_+(\widehat{D}_j^{(k)}, c_j)
  + \kl^{-1}_+(\widehat{\Delta}_j^{\,\mathcal{U}}, c_m)
\bigr]$.
\end{theorem}

\tcblower
\textbf{Interpretation}
{\small
\begin{itemize}[nosep,leftmargin=1.2em]
  \item[>]
  \ref{q:decode}\; $\widehat{R}_j^{(k)}$ small
     $\Longleftrightarrow$\; the representation has clean local class
      geometry.
  \item[>] \ref{q:agree}\; $\widehat{D}_j^{(k)}$ small
     $\Longleftrightarrow$\; the network computes through this
     representation.
  \item[>] \ref{q:stable}\; $\Delta_j^{\mathrm{stab}}$ small
     $\Longleftrightarrow$\; the training algorithm is stable
      under fold removal.
\end{itemize}}

\end{tcolorbox}

\begin{proof}
    See \cref{appx_proof_full} for the proof.
\end{proof}

The representation $\bm\varphi$
and metric $d$ enter the bound only through the three empirical quantities $\widehat{R}_j^{(k)}$, $\widehat{D}_j^{(k)}$, and
$\widehat{\Delta}_j^{\,\mathcal{U}}$; no parameter-space complexity measure appears anywhere. What makes the bound tight is not the proof technique but the learned representation, which creates a local class
geometry so clean that a trivial classifier nearly matches the network.
The bound measures that quality; it does not create it.

\section{Results}
\label{sec:experiments}

We evaluate the certificate in two settings: transfer with zero
target training (\cref{sec:transfer}) and cross-fit certification of
networks trained from scratch (\cref{sec:crossfit}), then compare
with prior bounds.
All bounds use $\ell_2$ distance on the last-hidden-layer
representation $h_{\bm\theta}\colon \bm X \to \mathbb{R}^D$,
the vector fed into the final linear classifier, computed by a
single unmodified forward pass.
For ResNets, $h_{\bm\theta}$ is the output of global average
pooling ($D{=}512$ or $2048$);
for Vision Transformers, the class token after the final layer norm
($D{=}768$ or $1024$);
for DINOv2, the class-token embedding from the frozen backbone.
Full extraction details are in \cref{appx:repr_details}.
\subsection{Transfer certification without training}
\label{sec:transfer}

When the feature map is fixed independently of the target data (e.g.\ pretrained
on ImageNet), \ref{q:agree} and \ref{q:stable} vanish and
\cref{prop:transfer} applies directly. The certified object is the
$k$-NN decoder built on top of the frozen representation, not the pretrained
network's own classifier head. We evaluate eight pretrained models plus a randomly initialized ResNet-18 as a
null control. For each, we extract the $\ell_2$ representation with a single
forward pass, split the target data 80/20 ($n_e {\approx}\, 10$k), and report
the tightest bound over $k \in \{1,3,5,7,11,15,21\}$ at $\delta {=} 0.05$ (the
reported certificates are best-of-grid; the union bound over $k$ is already
included). Four target datasets are used: CIFAR-10, CIFAR-100, ImageNet
validation, and ImageNet-V2 (distribution shift). No model undergoes any
target-task network training. \Cref{tab:transfer} reports the results.

\begin{table}[htb!]
\centering
\resizebox{12.5cm}{!}{%
\begin{tabular}{@{}ll rrr rrr rrr rrr@{}}
\toprule
& &
\multicolumn{3}{c}{CIFAR-10} &
\multicolumn{3}{c}{CIFAR-100} &
\multicolumn{3}{c}{ImageNet} &
\multicolumn{3}{c}{ImageNet-V2} \\
\cmidrule(lr){3-5}\cmidrule(lr){6-8}\cmidrule(lr){9-11}\cmidrule(lr){12-14}
Model & Params
  & Err & Bnd & Gap
  & Err & Bnd & Gap
  & Err & Bnd & Gap
  & Err & Bnd & Gap \\
\midrule
DINOv2-L/14  & 304M
  &  0.9 & \textbf{1.4} & \btight{0.5}
  & 10.4 & \textbf{12.1} & \btight{1.7}
  & 20.9 & 23.1 & \btight{2.2}
  & 37.5 & 43.0 & \bgood{5.5} \\
DINOv2-g/14  & 1.1B
  & --- & --- & ---
  & --- & --- & ---
  & 21.2 & 23.5 & \btight{2.3}
  & 37.7 & 43.2 & \bgood{5.5} \\
DINOv2-B/14  & 86M
  &  2.3 &  3.2 & \btight{0.9}
  & 14.9 & 16.9 & \btight{2.0}
  & 23.4 & 25.7 & \btight{2.3}
  & 41.3 & 46.8 & \bgood{5.5} \\
ViT-B/16     & 86.6M
  &  6.8 &  8.3 & \btight{1.5}
  & 28.1 & 30.5 & \btight{2.4}
  & 21.6 & 23.8 & \btight{2.2}
  & 39.2 & 44.7 & \bgood{5.5} \\
ViT-L/16     & 304M
  &  7.7 &  9.2 & \btight{1.5}
  & 29.5 & 32.0 & \btight{2.5}
  & 21.8 & 24.0 & \btight{2.2}
  & 40.3 & 45.8 & \bgood{5.5} \\
WRN-50-2     & 68.9M
  & 10.1 & 11.8 & \btight{1.7}
  & 34.3 & 36.9 & \btight{2.6}
  & 24.1 & 26.4 & \btight{2.3}
  & 47.6 & 53.1 & \bgood{5.5} \\
ResNet-50    & 25.6M
  & 12.1 & 13.9 & \btight{1.8}
  & 36.7 & 39.3 & \btight{2.6}
  & 26.5 & 28.9 & \btight{2.4}
  & 49.4 & 55.0 & \bgood{5.6} \\
ResNet-18    & 11.7M
  & 18.0 & 20.1 & \btight{2.1}
  & 43.6 & 46.2 & \btight{2.6}
  & 44.1 & 46.8 & \btight{2.7}
  & 70.2 & 75.1 & \btight{4.9} \\
\midrule
\textit{ResNet-18 (rand.)} & \textit{11.7M}
  & \textit{63.3} & \textit{65.8} & \btight{2.5}
  & \textit{87.8} & \textit{89.5} & \btight{1.7}
  & \textit{99.0} & \textit{99.4} & \btight{0.4}
  & \textit{99.5} & \textit{99.9} & \btight{0.4} \\
\bottomrule
\end{tabular}}
\caption{%
  \textbf{Transfer decoder certification ($\ell_2$, $\delta{=}0.05$).}
  The certified object is the $k$-NN decoder, not the pretrained
  classifier head (\ref{q:agree}, \ref{q:stable} vanish by
  construction). Err: $k$-NN error~(\%). Bnd: certified bound~(\%).
  Gap:
  {\scriptsize\colorbox{btight!25}{\strut$<$5}}
  {\scriptsize\colorbox{bgood!35}{\strut5--15}}
  {\scriptsize\colorbox{bok!40}{\strut15--30}}
  {\scriptsize\colorbox{bloose!40}{\strut30--50}}
  {\scriptsize\colorbox{bvac!30}{\strut$>$50\,pp}}.
}
\label{tab:transfer}
\end{table}

Three findings stand out. \textbf{(i)} First, the bounds are tight: DINOv2-L/14 certifies at
$1.4\%$ on CIFAR-10 and $12.1\%$ on CIFAR-100; all eight pretrained models
produce non-vacuous certificates on 1{,}000-class ImageNet, the tightest being
$23.1\%$. To our knowledge, these are the first non-vacuous ImageNet-scale certificates for $k$-NN decoders built from unmodified deterministic pretrained representations. \textbf{(ii)} Second, the certificate extends to
distribution shift: all eight models also certify on
ImageNet-V2~\citep{recht}, with the wider gap (${\sim}5.5$\,pp
vs.\ ${\sim}2.2$\,pp in-distribution) reflecting the harder setting rather than
a limitation of the bound. \textbf{(iii)} Third, the bound tracks representation quality, not
model size: DINOv2-g (1.1B params) and DINOv2-L (304M) produce nearly identical
certificates on ImageNet ($23.5\%$ vs.\ $23.1\%$), while ResNet-18 pretrained
vs.\ random, same 11.7M parameters, spans $20.1\%$ to $65.8\%$ on CIFAR-10.
The certificate measures what the network learned, not how many parameters were
used to express it.

\subsection{Cross-fit certification of trained networks}
\label{sec:crossfit}
We now certify networks trained from scratch using the full
certificate (\cref{cor:full}) with $J{=}5$ folds,
$\delta{=}\delta'{=}0.025$, and $\mathcal{U}$ = standard test inputs.
\Cref{tab:comparison_merged} reports the results alongside recent
certificates from prior work.

On MNIST all architectures certify tightly (best:
ResNet-18 at $2.6\%$); on CIFAR-10, ResNet-50 certifies at $12.4\%$; and on CIFAR-100, ResNet-50 certifies at $44.6\%$. To our knowledge, these are the first non-vacuous CIFAR-100 certificates we are aware of, and among the tightest CIFAR-10 certificates, for networks trained from scratch without modifying the certified network. The three-term decomposition shows \emph{why}. ResNet-50 on
CIFAR-10 splits as \ref{q:decode}~$=5.7\%$, \ref{q:agree}~$=1.8\%$,
\ref{q:stable}~$=1.2\%$: clean geometry, faithful decoder, stable
training. ViT on CIFAR-100 is vacuous and every term is large
(\ref{q:decode}~$=43.7\%$, \ref{q:agree}~$=20.2\%$,
\ref{q:stable}~$=35.4\%$). A small ViT on $40{,}000$ images does not
learn a stable, class-coherent representation. The same architecture
\emph{pretrained} certifies at $30.5\%$ (\cref{tab:transfer}), with
\ref{q:stable}${=}0$ by construction: same theorem, only the
representation changed. The bound diagnoses representation quality, not
architecture: FFNNs certify tightly on MNIST ($4.8\%$) and correctly
vacuously on CIFAR ($>95\%$, where they genuinely fail at $44\%$ test
error).

\begin{table}[htb!]
\centering
\resizebox{12.5cm}{!}{%
\begin{tabular}{@{}ll c rrr l@{}}
\toprule
Method & Architecture & $n$ & Test & Bound & Gap & Certified object \\
\midrule
\multicolumn{7}{@{}l}{\textit{\textbf{MNIST}}} \\[1pt]
\cite{dziugaite2017computingnonvacuousgeneralizationbounds}
  & FCN-600 & 60k & ${\sim}$2.5\% & ${\sim}$17\% & \bok{14.5} & Stochastic net \\
\cite{zhou2019nonvacuousgeneralizationboundsimagenet}
  & LeNet-5 & 60k & ${\sim}$1\% & ${\sim}$46\% & \bvac{45} & Compressed net \\
\cite{lotfi2022pacbayescompressionboundstight}
  & LeNet-5 & 60k & ${\sim}$1\% & 11.6\% & \bgood{10.6} & Quant.\ subspace \\
\cite{pérezortiz2021tighterriskcertificatesneural}
  & FCN & 60k & ${\sim}$1.4\% & ${\sim}$2.3\% & \btight{0.9} & Stochastic net \\
\addlinespace[3pt]
\hline
\addlinespace[3pt]
\rowcolor{gray!10} \textbf{Ours (FFNN)}
  & FFNN & 60k & 1.5\% & 4.9\% & \btight{3.4} & \textbf{Det.\ net via repr.} \\
\rowcolor{gray!10} \textbf{Ours (ViT)}
  & ViT-small & 60k & 0.9\% & 3.5\% & \btight{2.6} & \textbf{Det.\ net via repr.} \\
\rowcolor{gray!15} \textbf{Ours (ResNet)}
  & ResNet-18 & 60k & \textbf{0.6\%} & \textbf{2.6\%} & \btight{\textbf{2.0}} & \textbf{Det.\ net via repr.} \\
\midrule
\multicolumn{7}{@{}l}{\textit{\textbf{CIFAR-10}}} \\[1pt]
\cite{lotfi2022pacbayescompressionboundstight}
  & ConvNet & 50k & -- & ${\sim}$58.2\% & \bvac{--} & Quant.\ subspace \\
\cite{bazinet2026bounddisagreegeneralizationbounds}
  & ResNet-18 & 50k & 5.8\% & 17.36\% & \bok{11.5} & Det.\ net via surr. \\
\addlinespace[3pt]
\hline
\addlinespace[3pt]
\rowcolor{gray!10} \textbf{Ours (FFNN)}
  & FFNN & 50k & 44.6\% & 95.5\% & \bvac{50.9} & \textbf{Det.\ net via repr.} \\
\rowcolor{gray!10} \textbf{Ours (ViT)}
  & ViT-small & 50k & 12.3\% & 31.9\% & \bok{19.6} & \textbf{Det.\ net via repr.} \\
\rowcolor{gray!10} \textbf{Ours (ResNet-18)}
  & ResNet-18 & 50k & 4.7\% & 13.0\% & \bgood{8.3} & \textbf{Det.\ net via repr.} \\
\rowcolor{gray!15} \textbf{Ours (ResNet-50)}
  & ResNet-50 & 50k & \textbf{4.3\%} & \textbf{12.4\%} & \bgood{\textbf{8.0}} & \textbf{Det.\ net via repr.} \\
\midrule
\multicolumn{7}{@{}l}{\textit{\textbf{CIFAR-100}}} \\[1pt]
\cite{lotfi2022pacbayescompressionboundstight}
  & ConvNet & 50k & -- & 94.6\% & \bvac{--} & Quant.\ subspace \\
\addlinespace[3pt]
\hline
\addlinespace[3pt]
\rowcolor{gray!10} \textbf{Ours (FFNN)}
  & FFNN & 50k & 75.2\% & 100.0\% & \bvac{24.8} & \textbf{Det.\ net via repr.} \\
\rowcolor{gray!10} \textbf{Ours (ViT)}
  & ViT-small & 50k & 37.1\% & 99.3\% & \bvac{62.3} & \textbf{Det.\ net via repr.} \\
\rowcolor{gray!10} \textbf{Ours (ResNet-18)}
  & ResNet-18 & 50k & 21.2\% & 51.1\% & \bloose{29.9} & \textbf{Det.\ net via repr.} \\
\rowcolor{gray!15} \textbf{Ours (ResNet-50)}
  & ResNet-50 & 50k & \textbf{19.4\%} & \textbf{44.6\%} & \bok{\textbf{25.2}} & \textbf{Det.\ net via repr.} \\
\bottomrule
\end{tabular}}
\caption{\textbf{Cross-fit certification vs.\ prior bounds.}
Per-dataset blocks list prior work (above the rule) and our cross-fit
bounds across architecture families (below, shaded). Best row per dataset
in darker shade and bold. ReLU activation; SiLU results are in \Cref{tab:silu_comparison}. $\ell_2$ metric,
$\delta{=}\delta'{=}0.025$, best over $k \in \{1,3,5,7,11,15,21\}$ and
$j \in [5]$; union-bound cost included. Prior-work numbers from original
publications.}
\label{tab:comparison_merged}
\end{table}

\paragraph{Comparison with prior bounds.} On MNIST our cross-fit bound
($2.6\%$) is comparable to the tightest known
certificate~\citep{pérezortiz2021tighterriskcertificatesneural} while
certifying a deterministic network rather than a stochastic perturbation.
On CIFAR-10, our ResNet-18 bound ($13.0\%$) improves on
\citep{bazinet2026bounddisagreegeneralizationbounds} at the same
architecture ($17.36\%$); on CIFAR-100 we are not aware of prior
non-vacuous certificates. The comparison is not entirely apples-to-apples:
different methods certify different objects (stochastic nets, compressed
nets, surrogates, deterministic nets, decoders), catalogued in
\Cref{tab:desiderata}. The distinctive property of our approach is that
it certifies the deployed deterministic network without modified training
and extends to pretrained representations.



\section{What the bound reveals}
\label{sec:analysis}

\subsection{Overparameterization dissolves at the representation level}
\label{sec:overparam}

\begin{figure}[htb!]
\centering
\includegraphics[width=\linewidth]{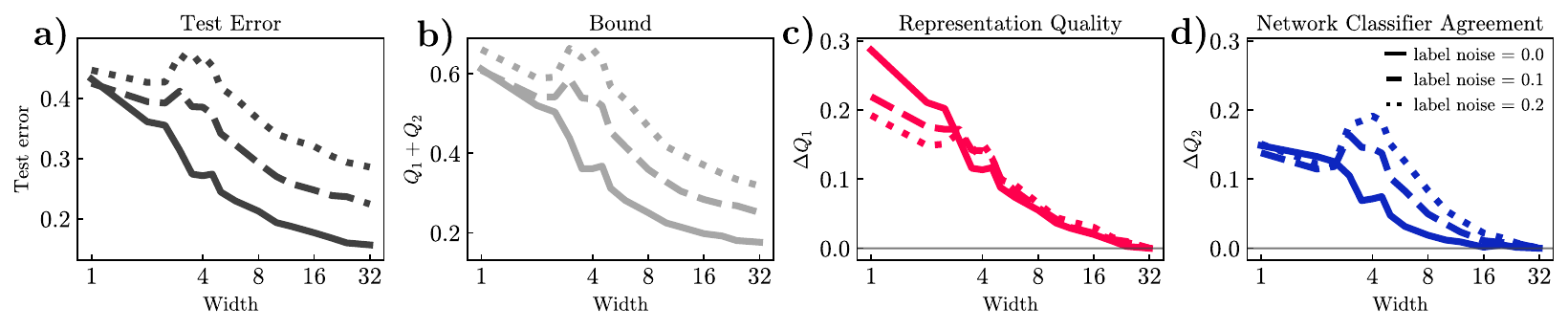}
\caption{ResNet-18 width sweep on CIFAR-10 (0\%, 10\%, 20\% label noise, 3 seeds). \textbf{a)} Test error and total bound. \textbf{b)} $\Delta Q_1(w)$ and $\Delta Q_2(w)$, each measured as excess over the widest-network baseline. \textbf{c)} $\Delta Q_1$ (local class geometry) decreases monotonically in width. \textbf{d)} $\Delta Q_2$ (decoder--network agreement) is the only non-monotone term, peaking at the interpolation threshold and then collapsing. The test-error hump is localized to $Q_2$. The $Q_3$ term is not analyzed in this sweep.}
\label{fig:width}
\end{figure}

Our bound does not index parameters, only the representation they produce. DINOv2-g (1.1B params) and DINOv2-L (304M) give nearly identical certificates on ImageNet ($23.5\%$ vs.\ $23.1\%$); ResNet-18 pretrained vs.\ random, same 11.7M parameters, spans $20.1\%$ to $65.8\%$ on CIFAR-10 (\cref{tab:transfer}). The relevant variable is representation quality, not parameter count. The width-scaling experiment (\cref{fig:width}) makes this precise and yields a sharper claim: the bound decomposes double descent into a monotone representation term and a transient head--representation misalignment. We train ResNet-18 at width multipliers ($w$) $1{\times}$--$32{\times}$ on CIFAR-10 with $0\%$, $10\%$, and $20\%$ label noise, and evaluate each term at every width. To isolate the shape of each term we report its excess over the widest-network baseline, $\Delta Q_1(w) = Q_1(w) - Q_1(w_{\max})$ and analogously for $\Delta Q_2$. $\Delta Q_1$ is monotone in width: wider networks produce strictly cleaner local class geometry, with label noise shifting the curve upward without changing its shape. $\Delta Q_2$ is the only non-monotone term, peaking sharply near the interpolation threshold and then collapsing. At the critical width the representation already supports accurate local classification, but the linear head fails to exploit this geometry. The classical double-descent curve in test error is therefore the sum of a monotone improvement in representation quality ($Q_1$) and a transient spike in head--representation misalignment ($Q_2$), both of which the certificate measures directly; the test-error hump is localized to $Q_2$. The overparameterization question carries an implicit assumption that parameter count is the relevant variable; our decomposition reframes it as: \emph{why does training on natural data produce representations where simple local classifiers succeed?}

\subsection{Local class geometry is causal, not correlational}
\label{sec:causal}

\begin{figure}[htb!]
  \centering
  \includegraphics[width=\linewidth]{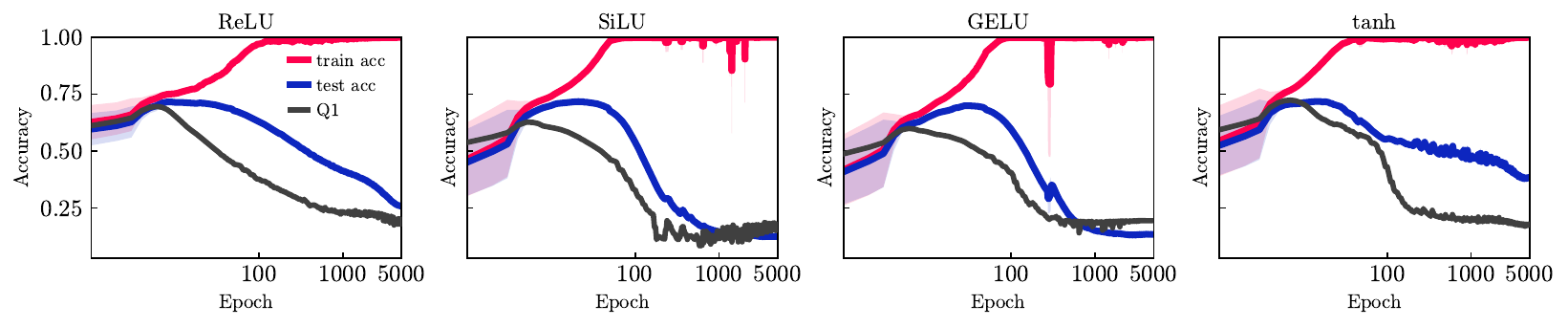}
  \caption{Six-layer FFNNs (width${=}1024$) trained on SVHN with $\mathcal{L}_{\mathrm{CE}} + \lambda\,\mathcal{L}_{\mathrm{rep}}$ across four activation functions. Training accuracy (\textcolor{blue}{blue}) saturates near $100\%$; test accuracy (\textcolor{red}{red}) degrades toward chance ($\sim 10\%$). $Q_1$ (\textcolor{black}{black}) decays together with test error in every case. The qualitative $Q_1$-to-test-error coupling is invariant across activation functions, although trajectories differ quantitatively. }
  \label{fig:causal_reg}
\end{figure}

The experiments above show that the bound \emph{tracks} generalization. Does it measure a cause, or a symptom? A natural worry is that $Q_1$ and generalization are both downstream of some third factor (good optimization or a well-conditioned loss landscape, for example), in which case $Q_1$ would be a correlate rather than a mechanism. We test this with an intervention that alters $Q_1$ while leaving the network's predictions on the training set unconstrained, and ask whether generalization follows. Define a regularizer targeting exactly what $Q_1$ measures, within-class representation similarity:
\begin{equation}
  \mathcal{L}_{\mathrm{rep}} \;=\;
  \frac{1}{|\mathcal{P}|}
  \sum_{(i,j)\in\mathcal{P}}
  \Bigl(\bar{h}(\bm{x}_i)^\top \bar{h}(\bm{x}_j)\Bigr)^{\!2},
  \qquad
  \mathcal{P} = \bigl\{(i,j): i\neq j,\; y_i = y_j,\; i,j\in S\bigr\},
  \label{eq:rep_reg}
\end{equation}
where $\bar{h}(\bm{x}) = h(\bm{x})/(\|h(\bm{x})\|_2 + \varepsilon)$ is the $\ell_2$-normalized representation at each hidden layer. Minimizing $\mathcal{L}_{\mathrm{rep}}$ pushes same-class representations apart, destroying the local class geometry that $Q_1$ quantifies. Crucially, it places no constraint on the network's outputs: with enough capacity the network can reach $\mathcal{L}_{\mathrm{rep}} \approx 0$ and $\mathcal{L}_{\mathrm{CE}} = 0$ simultaneously, i.e.\ perfect training accuracy with destroyed class geometry, as shown ing \Cref{fig:causal_reg}. We train six-layer FFNNs (width${=}1024$) on SVHN with $\mathcal{L}_{\mathrm{CE}} + \lambda\,\mathcal{L}_{\mathrm{rep}}$ across four activation functions. Training accuracy saturates near $100\%$ in every case: the network memorizes the data perfectly. Test accuracy strongly degrades. The trajectories differ across ReLU, SiLU, GELU, and tanh in both timing and the relative position of $Q_1$ and test error. But,  the qualitative pattern is invariant: training stays at $100\%$, while test accuracy collapses as the $Q_1$ deteriorates.  These results provide novel interventional evidence linking a certifiable quantity to generalization. By introducing a penalty that degrades the quality measure $Q_1$ without constraining the network's training predictions, we demonstrate that a generalizing network can be driven into a memorizing regime. Because this effect persists across the distinct optimization dynamics induced by four different activations, it strongly suggests that local class geometry acts as a fundamental mechanism separating generalization from mere memorization.


\subsection{Why parameter-space certificates can be loose}
\label{sec:pac-bayes}

\begin{figure}[htb]
  \centering
  \includegraphics[width=\linewidth]{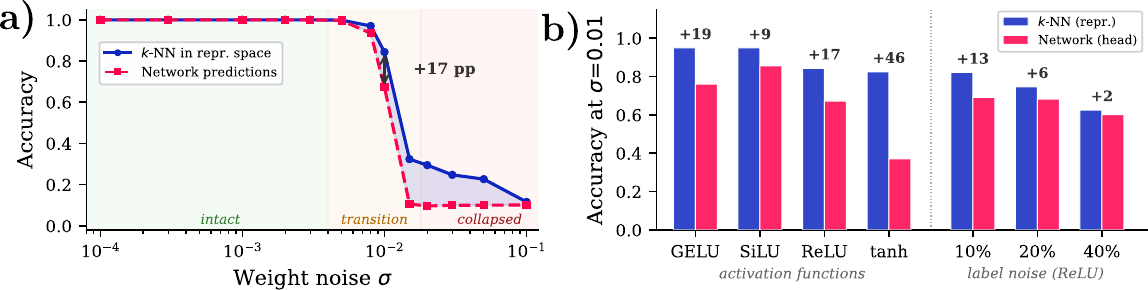}
  \caption{\textbf{a)}~ReLU ResNet-18 on CIFAR-10.  Under Gaussian weight noise at
    level~$\sigma$, $k$-NN accuracy in $\ell_2$ representation space (\textcolor{blue}{solid
    blue}) degrades more slowly than the network's own predictions (\textcolor{red}{dashed red}).
    At $\sigma{=}0.01$ the $k$-NN decoder retains $84\%$ accuracy while the
   network has fallen to $67\%$.
    \textbf{b)}~The representation advantage holds across activation functions
    and label-corruption levels at $\sigma{=}0.01$, with the gap ranging from
    $+2$\,pp (40\% corruption) to $+46$\,pp (tanh).}
  \label{fig:repr-stability}
  \vspace{-4mm}
\end{figure}

PAC-Bayes bounds certify a stochastic classifier $\hat
f_{\theta+\varepsilon}$ by adding noise to the weights. The bound is tight when
the full network, including the linear readout, tolerates perturbation.
\Cref{fig:repr-stability} suggests why this can be difficult: under Gaussian
weight noise, $k$-NN accuracy in representation space degrades far more slowly
than the network's own predictions. At $\sigma{=}0.01$ on a ReLU ResNet-18, the
$k$-NN decoder retains $84\%$ accuracy while the network has fallen to $67\%$.
The pattern holds across activations and label-corruption levels
(\cref{fig:repr-stability}b), with the gap ranging from $+2$\,pp to $+46$\,pp.
The linear readout is consistently the more fragile component. This does not
show that PAC-Bayes bounds cannot be tight; tighter variants
exist~\citep{pérezortiz2021tighterriskcertificatesneural,lotfi2022pacbayescompressionboundstight}. It
offers an intuition for why certifying through parameters can pay for the
fragility of the head, while certifying through the representation avoids this
cost.

\section{Discussion}
\label{sec:discussion}


\begin{tcolorbox}[colback=gray!5, colframe=black, boxrule=0.1pt,boxsep=3pt,     
    left=2pt,       
    right=2pt,      
    top=2pt,        
        bottom=2pt   ]   
The question we study is not:  \textbf{Why do overparameterized networks generalize?}, but rather: \textbf{What determines good generalisation at the representation level?}. This is a question about representation learning and data structure, not capacity control.
We find that good classification follows from clean local class geometry via elementary
concentration. Our bound provides a quantitative tool for measuring this:  a network with $10^8$ parameters and one with $10^5$ that induce the same feature geometry will receive the same bound.
\end{tcolorbox}

\paragraph{Limitations.} \ref{q:stable} is certified empirically rather than analytically. This ensures the bound remains usable across any optimizer (e.g., Adam), schedule, or augmentation without requiring re-derivation. The certificate specifically bounds classification error using last-layer $\ell_2$ distance; exploring calibration, robustness, regression, or alternate metrics is left to future work. Finally, cross-fit certification requires training $J$ fold models, a nontrivial one-off cost at ImageNet scale.

\paragraph{Broader impacts.} While our certificates enable transparent evaluation by localizing generalization bottlenecks, the primary risk is over-trust. Our guarantee strictly bounds classification error under stated assumptions; it does not certify fairness, privacy, robustness, or overall deployment safety. Therefore, these bounds should inform, rather than unilaterally approve, high-stakes systems.

\paragraph{Future work.} While our representation-level bound successfully decouples generalization from parameter-space capacity, it leaves open a crucial question: how does SGD reliably build this clean local class geometry in the first place? A natural next step is to investigate this mechanism through the emerging lens of ``learning mechanics'' \citep{simon2026scientifictheorydeeplearning}, exploring whether the formation of certifiable class geometry can be derived as a predictable, macroscopic observable governed by the fundamental dynamics of the training process.

\newpage
\begin{ack}
This work was supported by the Laboratory for AI Security Research (LASR). The views expressed in this paper are those of the authors and do not necessarily reflect the position of LASR or His Majesty’s Government.
\end{ack}

\bibliographystyle{plainnat}
\bibliography{bib}

\newpage


\appendix

\section{Appendix / supplemental material}


\section{Related work}
\label{appx_relwork}

While the history of generalization theory spans VC and uniform-convergence
analyses, PAC learning, Rademacher complexity, margin and norm-based bounds,
algorithmic stability, PAC-Bayes, compression, and modern non-vacuous
deep-network certificates~\citep{vapnik2015uniform,bartlett2002rademacher,
valiant1984theory,blumer1989learnability,bartlett1998boosting,
bartlett2002sample,mcallester1998some,hardt2016train,
neyshabur2015norm,bartlett2017spectrally,arora2018stronger,dziugaite2017computingnonvacuousgeneralizationbounds},
we focus on a smaller
universe: computable numerical certificates for a fixed trained predictor. In
that narrower sense, our proof template is closer to the classical hold-out
tradition than to global capacity measures. Practical prediction theory has long
emphasized that fresh-sample error, combined with a sharp concentration
inequality, already yields useful guarantees \citep{langford2005tutorial}. The
same perspective extends naturally to $K$-fold and progressive validation
\citep{blum1999beating}, leave-one-out and algorithmic-stability analyses
\citep{kearns1997algorithmic,bousquet2002stability}, and modern SGD-based
stability bounds \citep{hardt2016train}. Our cross-fit analysis is squarely in
this lineage: the mathematics is intentionally elementary, relying on held-out
Bernoulli errors, cross-fitting, and KL concentration. The unusual step is the
object being evaluated---a simple $k$-NN rule in the learned representation
\citep{cover1967nearest}.

The strongest computable non-vacuous certificates for deep networks have mostly
come from PAC-Bayes and compression. Classical PAC-Bayes work established
general-purpose risk bounds for randomized predictors
\citep{mcallester1998some,seeger2001improved,seeger2002pac}, and modern
deep-learning instantiations include spectrally normalized or localized
certificates with increasingly tight numerical performance
\citep{neyshabur2017pac,dziugaite2017computingnonvacuousgeneralizationbounds,pérezortiz2021tighterriskcertificatesneural}.
A parallel line certifies models through description length, pruning, or
compression
\citep{hinton1993keeping,arora2018stronger,zhou2019nonvacuousgeneralizationboundsimagenet,lotfi2022pacbayescompressionboundstight}.
These methods are the natural numerical baselines for any new certificate, but
they typically reason through parameter-space perturbation, posterior
complexity, or compressibility. A positioning nuance important here
is that PAC-Bayes is not only a parameter-space story: function-space and
marginal-likelihood PAC-Bayes analyses also exist
\citep{valle2020generalization,zhang2025position}. Our contribution is therefore
not ``the first non-parameter-space bound'' in the broadest sense, but a
particularly simple representation-space certificate based on held-out error,
agreement, and stability.

A smaller recent literature has moved closer to certifying the exact deployed
deterministic predictor. \citep{nagarajan2019deterministic} derive deterministic
PAC-Bayes bounds via noise resilience, and \citep{clerico2022generalisation}
study deterministic PAC-Bayes bounds along gradient-descent trajectories.
\citep{than2025nonvacuousgeneralizationboundsdeep} obtain non-vacuous bounds for unmodified trained
networks through an a priori data partition rather than post-hoc stochastic
perturbation or compression. Our results belong to this deterministic family,
but the route is different: instead of certifying the predictor through weight
perturbations, trajectory quantities, or partition structure alone, we certify a
simple representation-space decoder and then bridge from decoder to the deployed
network.

A particularly close modern comparator is disagreement-based certification.
PAC-Bayesian majority-vote analyses already show that disagreement can be a
certificate-relevant quantity \citep{germain2015risk}, and
\citep{bazinet2026bounddisagreegeneralizationbounds} turn this into a general recipe: certify an auxiliary
predictor, then use unlabeled disagreement to transfer the guarantee to the
target model. Our method is spiritually close to that template, but with less
auxiliary machinery. The surrogate is not separately trained; it is a
parameter-free $k$-NN decoder induced by the representation itself. The
unlabeled term is also more interpretable: it measures fold/full prediction
stability, which reconnects the argument to the classical stability literature
\citep{bousquet2002stability,hardt2016train}.

Outside the certification literature, the premise that a frozen
representation should be judged by simple probes is standard in modern vision.
Nearest-neighbor evaluation appears already in non-parametric instance
discrimination \citep{wu2018unsupervised} and remains common, alongside linear
probing, in transfer and self-supervised learning
\citep{kornblith2019better,kolesnikov2020big,chen2020simple,caron2021emerging,oquab2023dinov2}.
Relatedly, \citet{ding2022pactran} use PAC-Bayesian quantities to estimate the
transferability of pretrained representations. Our contribution is therefore not
the observation that $k$-NN is a useful probe. It is that, with cross-fitting,
agreement, and stability, held-out probe performance becomes a formal
certificate for the deployed network itself.

The empirical claims of this paper also touch two further literatures. First, modern work on double descent studies the
interpolation peak in model size, training time, and sample size
\citep{belkin2019reconciling,nakkiran2021deep}. Second, neural-collapse work
studies the simple class geometry that emerges in the last layer of trained
classifiers, including nearest-class-center behavior
\citep{papyan2020prevalence}. Our \ref{q:decode} term is not a neural-collapse
theorem, but it measures a nearby object: local class geometry sufficient for a
simple local classifier to succeed. Likewise, our causal manipulation of
\ref{q:decode} should be read against memorization results showing that deep
networks can fit labels without learning useful structure
\citep{zhang2017rethinking}. Finally, unlike post-hoc complexity measures that
are primarily judged by correlation with test error across model families
\citep{jiang2020fantastic}, our method returns a finite-sample upper bound.

\section{Further results}
\subsection{SiLU comparison}

\begin{table}[htb!]
\centering
\resizebox{11cm}{!}{%
\begin{tabular}{@{}ll rrr rrr rrr@{}}
\toprule
 & & \multicolumn{3}{c}{MNIST ($n{=}60$k)}
   & \multicolumn{3}{c}{CIFAR-10 ($n{=}50$k)}
   & \multicolumn{3}{c}{CIFAR-100 ($n{=}50$k)} \\
\cmidrule(lr){3-5}\cmidrule(lr){6-8}\cmidrule(lr){9-11}
Model & Act.
  & Test & Bound & Gap
  & Test & Bound & Gap
  & Test & Bound & Gap \\
\midrule
\multicolumn{11}{l}{\textit{Feedforward networks}} \\[2pt]
FFNN & ReLU
  & 1.5 & 4.9{\tiny $\pm$0.4} & \btight{3.4}
  & 44.6 & 95.5{\tiny $\pm$1.1} & \bvac{50.9}
  & 75.2 & 100.0 & \bvac{24.8} \\
FFNN & SiLU
  & 1.4 & 4.8{\tiny $\pm$0.1} & \btight{3.4}
  & 44.3 & 96.2{\tiny $\pm$1.0} & \bvac{51.9}
  & 75.4 & 100.0 & \bvac{24.6} \\
\midrule
\multicolumn{11}{l}{\textit{Convolutional networks}} \\[2pt]
ResNet-18 & ReLU
  & 0.6 & \textbf{2.6}{\tiny $\pm$0.0} & \btight{2.0}
  & 4.7 & 13.0{\tiny $\pm$0.3} & \bgood{8.3}
  & 21.2 & 51.1{\tiny $\pm$0.1} & \bloose{29.9} \\
ResNet-18 & SiLU
  & 0.7 & 2.7{\tiny $\pm$0.1} & \btight{2.0}
  & 5.5 & 15.4{\tiny $\pm$0.1} & \bgood{9.9}
  & 23.0 & 58.6{\tiny $\pm$0.2} & \bloose{35.6} \\[3pt]
\rowcolor{gray!8}
ResNet-50 & ReLU
  & 0.7 & 2.9{\tiny $\pm$0.2} & \btight{2.2}
  & \textbf{4.3} & \textbf{12.4}{\tiny $\pm$0.4} & \bgood{8.0}
  & \textbf{19.4} & \textbf{44.6}{\tiny $\pm$0.3} & \bok{25.2} \\
\rowcolor{gray!8}
ResNet-50 & SiLU
  & 0.9 & 3.4{\tiny $\pm$0.1} & \btight{2.5}
  & 5.3 & 14.3{\tiny $\pm$0.2} & \bgood{8.9}
  & 21.8 & 55.9{\tiny $\pm$0.5} & \bloose{34.0} \\
\midrule
\multicolumn{11}{l}{\textit{Vision Transformers (from scratch)}} \\[2pt]
ViT & ReLU
  & 0.9 & 3.5{\tiny $\pm$0.2} & \btight{2.6}
  & 12.3 & 31.9{\tiny $\pm$0.5} & \bok{19.6}
  & 37.1 & 99.3{\tiny $\pm$0.2} & \bvac{62.3} \\
ViT & SiLU
  & 0.9 & 3.7{\tiny $\pm$0.1} & \btight{2.8}
  & 12.3 & 32.3{\tiny $\pm$0.3} & \bok{20.1}
  & 38.2 & 100.0 & \bvac{61.8} \\
\bottomrule
\end{tabular}}
\caption{%
  \textbf{Cross-fit certification of trained networks}
  ($\ell_2$, $J{=}5$, $\delta{=}\delta'{=}0.025$).
  Best bound over $k \in \{1,3,5,7,11,15,21\}$ and $j \in [5]$.
  Each entry: mean $\pm$ std over 3 seeds.
}
\label{tab:silu_comparison}
\end{table}

\begin{table}[t]
\centering
\caption{Desiderata for generalization certificates. Our approach
satisfies all desiderata; the stability term is certified empirically
from unlabeled data rather than bounded analytically.}
\label{tab:desiderata}
\scriptsize
\setlength{\tabcolsep}{2pt}
\renewcommand{\arraystretch}{1.1}
\resizebox{\textwidth}{!}{%
\begin{tabular}{@{}lcccccccc@{}}
\toprule
Method
  & \shortstack{Det.\\net}
  & \shortstack{No mod.\\training}
  & \shortstack{No\\surrogate}
  & \shortstack{No\\compress.}
  & \shortstack{No stoch.\\weights}
  & \shortstack{Pretrained\\/ transfer}
  & \shortstack{Arch.-\\agnostic}
  & \shortstack{Diagnoses\\repr.} \\
\midrule
\citet{dziugaite2017computingnonvacuousgeneralizationbounds}
  & \cellcolor{bvac!30}\xmark & \cellcolor{bvac!30}\xmark
  & \cellcolor{btight!25}\cmark & \cellcolor{btight!25}\cmark
  & \cellcolor{bvac!30}\xmark & \cellcolor{bvac!30}\xmark
  & \cellcolor{btight!25}\cmark & \cellcolor{bvac!30}\xmark \\
\citet{pérezortiz2021tighterriskcertificatesneural}
  & \cellcolor{bvac!30}\xmark & \cellcolor{bvac!30}\xmark
  & \cellcolor{btight!25}\cmark & \cellcolor{btight!25}\cmark
  & \cellcolor{bvac!30}\xmark & \cellcolor{bvac!30}\xmark
  & \cellcolor{btight!25}\cmark & \cellcolor{bvac!30}\xmark \\
\citet{zhou2019nonvacuousgeneralizationboundsimagenet}
  & \cellcolor{bvac!30}\xmark & \cellcolor{bvac!30}\xmark
  & \cellcolor{btight!25}\cmark & \cellcolor{bvac!30}\xmark
  & \cellcolor{bvac!30}\xmark & \cellcolor{bvac!30}\xmark
  & \cellcolor{btight!25}\cmark & \cellcolor{bvac!30}\xmark \\
\citet{lotfi2022pacbayescompressionboundstight}
  & \cellcolor{bvac!30}\xmark & \cellcolor{bvac!30}\xmark
  & \cellcolor{btight!25}\cmark & \cellcolor{bvac!30}\xmark
  & \cellcolor{bvac!30}\xmark & \cellcolor{btight!25}\cmark
  & \cellcolor{btight!25}\cmark & \cellcolor{bok!40}$\sim$ \\
\citet{bazinet2026bounddisagreegeneralizationbounds}
  & \cellcolor{btight!25}\cmark & \cellcolor{bvac!30}\xmark
  & \cellcolor{bvac!30}\xmark & \cellcolor{bok!40}$\sim$
  & \cellcolor{btight!25}\cmark & \cellcolor{bvac!30}\xmark
  & \cellcolor{btight!25}\cmark & \cellcolor{bvac!30}\xmark \\
\citet{than2025nonvacuousgeneralizationboundsdeep}
  & \cellcolor{btight!25}\cmark & \cellcolor{btight!25}\cmark
  & \cellcolor{btight!25}\cmark & \cellcolor{btight!25}\cmark
  & \cellcolor{btight!25}\cmark & \cellcolor{btight!25}\cmark
  & \cellcolor{bvac!30}\xmark & \cellcolor{bvac!30}\xmark \\
\midrule
\textbf{Ours}
  & \cellcolor{btight!25}\cmark & \cellcolor{btight!25}\cmark
  & \cellcolor{btight!25}\cmark & \cellcolor{btight!25}\cmark
  & \cellcolor{btight!25}\cmark & \cellcolor{btight!25}\cmark
  & \cellcolor{btight!25}\cmark & \cellcolor{btight!25}\cmark \\
\bottomrule
\end{tabular}}
\end{table}

\newpage

\section{Proofs}

\subsection{Proof of \texorpdfstring{\cref{thm:decoder}}{Theorem~\ref{thm:decoder}}}
\label{appx:proof_decoder}
\begin{proof}
    Fix $j \in [J]$ and $k \in \mathcal{K}$.  Conditioned on the
complement sample $\mathcal{S}_{-j}$, the fold-$j$ parameters
$\bm\theta^{(-j)} = \mathcal{A}(\mathcal{S}_{-j})$ are fixed, so the
decoder $h_j^{(k)}$ is a deterministic function (the reference set and
the $k$-NN rule are fully determined by $\mathcal{S}_{-j}$ and the
pre-specified tie-breaking).  The fold points
$\{(\bm x_i, y_i)\}_{i \in F_j}$ are i.i.d.\ draws from~$P$ that are
conditionally independent of~$\mathcal{S}_{-j}$.

Define the conditional risk
\[
  R_j^{(k)}
  \;=\;
  \Pr_{(\bm X, Y) \sim P}\!\bigl(
    h_j^{(k)}(\bm X) \neq Y
    \mid \mathcal{S}_{-j}
  \bigr).
\]
The indicators
$Z_i = \mathbf{1}\bigl[h_j^{(k)}(\bm x_i) \neq y_i\bigr]$,
$i \in F_j$, are conditionally i.i.d.\
$\mathrm{Bernoulli}\bigl(R_j^{(k)}\bigr)$, and their empirical mean
is $\widehat{R}_j^{(k)} = \frac{1}{n_j}\sum_{i \in F_j} Z_i$.

We apply the KL concentration inequality for i.i.d.\ Bernoulli
random variables
\citep[see e.g.][Theorem~1]{maurer2004note}: for $n$ i.i.d.\
$\mathrm{Bernoulli}(p)$ variables with empirical mean~$\hat{p}$,
\[
  \Pr\!\bigl(\kl(\hat{p} \| p) > c\bigr)
  \;\le\;
  (n+1)\,e^{-nc}
  \qquad \text{for all } c > 0.
\]
Applied conditionally on $\mathcal{S}_{-j}$, with
$\delta' > 0$ to be chosen, this gives
\[
  \Pr\!\bigl(
    R_j^{(k)} > \kl^{-1}_+\!\bigl(
      \widehat{R}_j^{(k)},\;
      \tfrac{\log((n_j{+}1)/\delta')}{n_j}
    \bigr)
    \;\big|\; \mathcal{S}_{-j}
  \bigr)
  \;\le\; \delta'
  \qquad \text{a.s.}
\]
Define the event
\[
  E_{j,k}
  \;=\;
  \Bigl\{
    R_j^{(k)} \le \kl^{-1}_+\!\bigl(
      \widehat{R}_j^{(k)},\;
      \tfrac{\log((n_j{+}1)/\delta')}{n_j}
    \bigr)
  \Bigr\}.
\]
The conditional bound gives
$\Pr(E_{j,k}^c \mid \mathcal{S}_{-j}) \le \delta'$ a.s.,
so by the tower property:
\[
  \Pr(E_{j,k}^c)
  \;=\;
  \mathbb{E}\bigl[
    \Pr(E_{j,k}^c \mid \mathcal{S}_{-j})
  \bigr]
  \;\le\; \delta'.
\]
A union bound over all $|\mathcal{K}| \cdot J$ pairs $(k,j)$ gives
\[
  \Pr\!\Bigl(
    \exists\, j \in [J],\; k \in \mathcal{K}
    \;:\; E_{j,k}^c
  \Bigr)
  \;\le\;
  |\mathcal{K}| \cdot J \cdot \delta'.
\]
Setting $\delta' = \delta / (|\mathcal{K}| \cdot J)$ yields: with
probability at least $1 - \delta$ over $\mathcal{S}_n$, for all
$j \in [J]$ and $k \in \mathcal{K}$,
\[
  R_j^{(k)}
  \;\le\;
  \kl^{-1}_+\!\left(
    \widehat{R}_j^{(k)},\;\;
    \frac{
      \log\!\bigl(|\mathcal{K}|\,J / \delta\bigr)
      + \log(n_j {+} 1)
    }{n_j}
  \right),
\]
which is \cref{eq:decoder}.
\end{proof}

\subsection{Proof of \cref{prop:transfer}}
\label{appx:proof_2}

\begin{proof}
Since $\bm\varphi$ is fixed by assumption, conditioning on
$\mathcal{S}_{\mathrm{ref}}$ makes the decoder $h^{(k)}$
deterministic (the reference set fixes the $k$-NN rule).
The evaluation points
$\{(\bm x_i, y_i)\}_{i \in I_{\mathrm{eval}}}$ are
therefore conditionally i.i.d.\ from~$P$.

The error indicators
$Z_i^{(k)} = \mathbf{1}\bigl[h^{(k)}(\bm x_i) \neq y_i\bigr]$,
$i \in I_{\mathrm{eval}}$, are conditionally i.i.d.\
$\mathrm{Bernoulli}(R^{(k)})$, where
$R^{(k)} = \Pr(\,h^{(k)}(\bm X) \neq Y \mid
\mathcal{S}_{\mathrm{ref}})$.
The KL concentration inequality applied conditionally on
$\mathcal{S}_{\mathrm{ref}}$ gives, for each $k \in \mathcal{K}$:
\[
  \Pr\!\bigl(
    R^{(k)} > \kl^{-1}_+\!\bigl(
      \widehat{R}^{(k)},\;
      \tfrac{\log((n_e{+}1)/\delta')}{n_e}\bigr)
    \;\big|\; \mathcal{S}_{\mathrm{ref}}
  \bigr)
  \;\le\; \delta'
  \quad\text{a.s.}
\]
By the tower property, each conditional bound promotes to an
unconditional one (as in the proof of \cref{thm:decoder}).
A union bound over all $|\mathcal{K}|$ values of~$k$ gives total
failure probability at most $|\mathcal{K}| \cdot \delta'$.
Setting $\delta' = \delta/|\mathcal{K}|$ yields
\[
  c_e
  \;=\;
  \frac{
    \log\!\bigl(|\mathcal{K}|/\delta\bigr)
    + \log(n_e{+}1)
  }{n_e},
\]
which is~\eqref{eq:transfer-decoder}.
\end{proof}

\subsection{Proof of \cref{thm:network}}
\label{appx:proof_3}

\begin{proof}
The indicator decomposition
\[
  \mathbf{1}\bigl[\widehat{f}_j(\bm x) \neq y\bigr]
  \;\le\;
  \mathbf{1}\bigl[h_j^{(k)}(\bm x) \neq y\bigr]
  \;+\;
  \mathbf{1}\bigl[h_j^{(k)}(\bm x) \neq \widehat{f}_j(\bm x)\bigr]
\]
splits the fold network's risk into the population decoder error and
the population disagreement rate:
\[
  R_j^{(k)}
  = \Pr\!\bigl(h_j^{(k)}(\bm X) \neq Y \mid \mathcal{S}_{-j}\bigr),
  \qquad
  D_j^{(k)}
  = \Pr\!\bigl(h_j^{(k)}(\bm X) \neq \widehat{f}_j(\bm X)
    \mid \mathcal{S}_{-j}\bigr).
\]
Conditioned on $\mathcal{S}_{-j}$, both $h_j^{(k)}$ and
$\widehat{f}_j$ are deterministic, so the fold points
$\{(\bm x_i, y_i)\}_{i \in F_j}$ are conditionally i.i.d.\ from~$P$.
This produces two families of conditionally i.i.d.\ Bernoulli
indicators:
\[
  Z_i^{(k)} = \mathbf{1}\bigl[h_j^{(k)}(\bm x_i) \neq y_i\bigr]
  \sim \mathrm{Bernoulli}\bigl(R_j^{(k)}\bigr),
  \qquad
  W_i^{(k)} = \mathbf{1}\bigl[h_j^{(k)}(\bm x_i) \neq
    \widehat{f}_j(\bm x_i)\bigr]
  \sim \mathrm{Bernoulli}\bigl(D_j^{(k)}\bigr),
\]
with empirical means $\widehat{R}_j^{(k)}$ and
$\widehat{D}_j^{(k)}$ respectively.

There are $2 \cdot |\mathcal{K}| \cdot J$ events to control
(two families $\times$ $|\mathcal{K}|$ values of~$k$ $\times$ $J$
folds).  Assigning confidence budget
$\delta' = \delta / (2\,|\mathcal{K}|\,J)$ to each, the KL
concentration inequality applied conditionally on
$\mathcal{S}_{-j}$ gives
\[
  \Pr\!\bigl(
    R_j^{(k)} > \kl^{-1}_+\!\bigl(
      \widehat{R}_j^{(k)},\;
      \tfrac{\log((n_j{+}1)/\delta')}{n_j}\bigr)
    \;\big|\; \mathcal{S}_{-j}
  \bigr)
  \;\le\; \delta'
  \quad\text{a.s.,}
\]
and identically for $D_j^{(k)}$ with $\widehat{D}_j^{(k)}$.
By the tower property, each conditional bound promotes to an
unconditional one.  A union bound over all
$2\,|\mathcal{K}|\,J$ events gives total failure probability at
most~$\delta$.  Substituting
$\delta' = \delta/(2\,|\mathcal{K}|\,J)$ into the log yields
$c_j = \bigl[\log(2\,|\mathcal{K}|\,J/\delta)
  + \log(n_j{+}1)\bigr]/n_j$,
and the risk decomposition
$\Pr(\widehat{f}_j(\bm X) \neq Y \mid \mathcal{S}_{-j})
  \le R_j^{(k)} + D_j^{(k)}$
gives~\eqref{eq:network}.
\end{proof}

\subsection{Proof of \cref{prop:q3_unlabeled} }
\label{appx:proof_stab}
\begin{proof}
Conditioned on $\mathcal{S}_n$, both $\widehat{f}_j$ and
$\widehat{f}^{(\mathrm{full})}$ are deterministic.
The indicators
$V_r = \mathbf{1}[\widehat{f}_j(\bm x_r) \neq
  \widehat{f}^{(\mathrm{full})}(\bm x_r)]$, $r = 1, \dots, m$,
are therefore conditionally i.i.d.\
$\mathrm{Bernoulli}(\Delta_j^{\mathrm{stab}})$,
with empirical mean $\widehat{\Delta}_j^{\,\mathcal{U}}$.

Assign confidence budget $\delta'' = \delta'/J$ to each fold.
The KL concentration inequality applied conditionally on
$\mathcal{S}_n$ gives
\[
  \Pr\!\bigl(
    \Delta_j^{\mathrm{stab}} >
    \kl^{-1}_+\!\bigl(
      \widehat{\Delta}_j^{\,\mathcal{U}},\;
      \tfrac{\log((m{+}1)/\delta'')}{m}\bigr)
    \;\big|\; \mathcal{S}_n
  \bigr)
  \;\le\; \delta''
  \quad\text{a.s.}
\]
By the tower property,
$\Pr(\cdot) = \mathbb{E}[\Pr(\cdot \mid \mathcal{S}_n)]
  \le \delta''$.
A union bound over $j \in [J]$ gives total failure probability at
most $J \cdot \delta'' = \delta'$.
Substituting $\delta'' = \delta'/J$ into the log yields the
confidence radius
$\bigl[\log(J/\delta') + \log(m{+}1)\bigr]/m$,
which is~\eqref{eq:q3_cert}.
\end{proof}

\subsection{Proof of \cref{cor:full}}
\label{appx_proof_full}
\begin{proof}
The proof combines two high-probability events via a union bound.

\emph{Event~1} (probability $\ge 1 - \delta$ over $\mathcal{S}_n$).
\Cref{thm:network} gives, for all $j \in [J]$ and
$k \in \mathcal{K}$:
\[
  \Pr_{(\bm X, Y) \sim P}\!\bigl(
    \widehat{f}_j(\bm X) \neq Y \mid \mathcal{S}_{-j}\bigr)
  \;\le\;
  \kl^{-1}_+\!\bigl(\widehat{R}_j^{(k)}, c_j\bigr)
  + \kl^{-1}_+\!\bigl(\widehat{D}_j^{(k)}, c_j\bigr).
\]

\emph{Event~2} (probability $\ge 1 - \delta'$ over
$(\mathcal{S}_n, \mathcal{U})$).
\Cref{prop:q3_unlabeled} gives, for all $j \in [J]$:
\[
  \Delta_j^{\mathrm{stab}}
  = \Pr_{\bm X \sim P_X}\!\bigl(
    \widehat{f}_j(\bm X) \neq
    \widehat{f}^{(\mathrm{full})}(\bm X)
    \mid \mathcal{S}_n\bigr)
  \;\le\;
  \kl^{-1}_+\!\bigl(
    \widehat{\Delta}_j^{\,\mathcal{U}}, c_m\bigr).
\]

On the intersection of both events (probability
$\ge 1 - \delta - \delta'$), the indicator
decomposition~\eqref{eq:triangle_stab} gives
\[
  \Pr_{(\bm X, Y) \sim P}\!\bigl(
    \widehat{f}^{(\mathrm{full})}(\bm X) \neq Y
    \mid \mathcal{S}_n\bigr)
  \;\le\;
  \Pr_{(\bm X, Y) \sim P}\!\bigl(
    \widehat{f}_j(\bm X) \neq Y
    \mid \mathcal{S}_n\bigr)
  \;+\;
  \Delta_j^{\mathrm{stab}}.
\]
Since $\widehat{f}_j$ depends on $\mathcal{S}_n$ only through
$\mathcal{S}_{-j}$, the first term equals
$\Pr(\widehat{f}_j(\bm X) \neq Y \mid \mathcal{S}_{-j})$,
which is bounded by Event~1.  Substituting both bounds
yields~\eqref{eq:main}.
\end{proof}


\section{Experimental Details}
\label{app:experiments}

This appendix collects all information needed to reproduce the experiments.
Section~\ref{app:common} describes shared ingredients; Sections~\ref{app:table1}--\ref{app:table4} give per-table details; Sections~\ref{app:fig1b}--\ref{app:fig4} give per-figure details; Section~\ref{app:seeds-compute} records seeds and compute.

\subsection{Common setup}
\label{app:common}

\paragraph{Datasets.}
The supervised experiments use the standard train/test splits of MNIST, CIFAR-10, CIFAR-100, and SVHN.
The transfer experiments additionally use the ImageNet ILSVRC-2012 validation set and ImageNet-V2.
Evaluation is always deterministic and uses the standard held-out test or validation inputs.

\paragraph{Representation and decoder.}
The representation~$\varphi(x)$ is the final hidden feature vector immediately before the linear readout: the output of global average pooling for convolutional networks, and the class-token embedding after the final layer norm for transformers.
The decoder is a majority-vote $k$-nearest-neighbor rule in $\ell_2$ representation space, searching over
$k\in\{1,3,5,7,11,15,21\}$.
When a single bound is reported it is always the minimum over this grid; the union bound over $|K|=7$ values of $k$ is included in the confidence term.
Distances are computed in \texttt{float64}; neighbor ties are broken by lowest reference index, and majority-vote ties by lowest class label.

\paragraph{Architectures.}
Table~\ref{tab:architectures} summarizes the four architecture families used in the trained-model experiments.
The transfer experiments use pretrained ResNet-18, ResNet-50, WRN-50-2, ViT-B/16, ViT-L/16, DINOv2-B/14, DINOv2-L/14, and DINOv2-g/14, plus a randomly initialized ResNet-18 as a null control (see Appendix~\ref{appx:repr_details} for extraction details).

\begin{table}[h]
\centering\small
\caption{Architecture details for trained-model experiments.}
\label{tab:architectures}
\begin{tabular}{llp{8.5cm}}
\toprule
Family & Repr.\ dim & Notes \\
\midrule
FFNN & $D{=}1024$ & 6 hidden layers, width 1024, no dropout. Activation (ReLU or SiLU) applied before readout. No data augmentation. \\
ResNet-18 & $D{=}512$ & CIFAR-style stem (3$\times$3 conv, no maxpool) for 32$\times$32 inputs. Standard torchvision stem for 224$\times$224. \\
ResNet-50 & $D{=}2048$ & Same stem conventions as ResNet-18. \\
ViT (from scratch) & varies & MNIST: patch 4, embed 192, depth 4, 3 heads, MLP ratio 4, dropout 0. CIFAR-10/100: patch 4, embed 256, depth 6, 8 heads, MLP ratio 4, dropout 0. \\
\bottomrule
\end{tabular}
\end{table}

\paragraph{Optimization recipes.}
Table~\ref{tab:hyperparams} lists every hyperparameter for the cross-fit experiments of Table~2. Within each architecture family the recipe is fixed across activations; differences across datasets are shown explicitly.

\begin{table}[h]
\centering\small
\caption{Optimization hyperparameters for the cross-fit experiments (Table~2). All schedules are cosine decay to zero unless stated otherwise. ``LS'' denotes label smoothing.}
\label{tab:hyperparams}
\begin{tabular}{ll ccccccc}
\toprule
Family & Dataset & Optimizer & LR & Momentum & WD & LS & Batch & Epochs \\
\midrule
\multirow{3}{*}{FFNN}
  & MNIST    & AdamW & $10^{-3}$          & --  & $10^{-5}$          & 0   & 512 & 150 \\
  & CIFAR-10 & AdamW & $5{\cdot}10^{-4}$  & --  & $2{\cdot}10^{-5}$  & 0   & 512 & 300 \\
  & CIFAR-100& AdamW & $3{\cdot}10^{-4}$  & --  & $2{\cdot}10^{-5}$  & 0   & 512 & 400 \\
\midrule
\multirow{3}{*}{ResNet-18}
  & MNIST    & SGD   & 0.05               & 0.9 & $10^{-4}$          & 0   & 256 & 120 \\
  & CIFAR-10 & SGD   & 0.1                & 0.9 & $5{\cdot}10^{-4}$  & 0   & 128 & 200 \\
  & CIFAR-100& SGD   & 0.1                & 0.9 & $5{\cdot}10^{-4}$  & 0.1 & 128 & 240 \\
\midrule
\multirow{3}{*}{ResNet-50}
  & MNIST    & SGD   & 0.05               & 0.9 & $10^{-4}$          & 0   & 128 & 120 \\
  & CIFAR-10 & SGD   & 0.1                & 0.9 & $5{\cdot}10^{-4}$  & 0   & 128 & 240 \\
  & CIFAR-100& SGD   & 0.1                & 0.9 & $5{\cdot}10^{-4}$  & 0.1 & 128 & 300 \\
\midrule
\multirow{3}{*}{ViT}
  & MNIST    & AdamW & $5{\cdot}10^{-4}$  & --  & 0.01               & 0   & 256 & 200 \\
  & CIFAR-10 & AdamW & $5{\cdot}10^{-4}$  & --  & 0.05               & 0.1 & 128 & 300 \\
  & CIFAR-100& AdamW & $5{\cdot}10^{-4}$  & --  & 0.05               & 0.1 & 128 & 300 \\
\bottomrule
\end{tabular}
\vspace{2pt}
\begin{flushleft}
\footnotesize
ViT uses a 10-epoch linear warmup before cosine decay. CIFAR-scale convnets and ViTs use standard random-crop-and-flip augmentation during training; FFNNs use no augmentation. All AdamW runs use default $(\beta_1,\beta_2)=(0.9,0.999)$.
\end{flushleft}
\end{table}

\subsection{Cross-fit construction}
\label{app:crossfit}

The cross-fit procedure is shared by Table~2, Figure~1b, and Figure~2.
We describe it once here; later subsections refer back.

\paragraph{Fold partition.}
The $n$ training indices are permuted with a fixed fold seed ($42$) and divided into $J{=}5$ contiguous blocks.
The partition is a single uniform random permutation; it is \emph{not} class-stratified.
For the balanced datasets used here, a five-way random split still yields large per-class counts in every fold (about 1\,200 per class for MNIST, 1\,000 for CIFAR-10, 100 for CIFAR-100), so we accept the realized folds without imposing balance.

\paragraph{Fold-model training.}
Each fold model is trained on the $0.8n$ complement examples using the \emph{same} architecture, hyperparameters, and number of epochs as the full-data model.
The batch size is unchanged, so a fold model executes fewer gradient steps per epoch than the full model.
We intentionally do not compensate for this by rescaling epochs or learning rates: the Q3 term is designed to measure the stability of the \emph{entire training recipe} under fold removal, including any optimization-trajectory change induced by the smaller sample.

\paragraph{Certification computation.}
For each fold~$j$ and each $k$, we compute three quantities on the held-out fold and the independent unlabeled test set:
\begin{enumerate}[leftmargin=2em,itemsep=0pt]
  \item \textbf{Q1} (decoder error): fraction of held-out fold examples where the $k$-NN decoder disagrees with the true label.
  \item \textbf{Q2} (decoder--network disagreement): fraction of the same held-out examples where the decoder disagrees with the fold-trained network.
  \item \textbf{Q3} (fold--full stability): fraction of independent test inputs where the fold-trained and full-data networks disagree. Labels are not used.
\end{enumerate}
Each empirical rate is converted to a KL upper bound with confidence radii
\[
  c_j = \frac{\log(2|K|\,J/\delta)+\log(|F_j|{+}1)}{|F_j|},
  \qquad
  c_m = \frac{\log(J/\delta')+\log(m{+}1)}{m},
\]
using $\delta{=}\delta'{=}0.025$.
The three bounds are summed, and the reported certificate is $\min_{j,k}$ of that sum.
The optimization over $(j,k)$ is therefore part of the stated certificate, not an unaccounted model-selection step.


\subsection{Table~1: Transfer certification}
\label{app:table1}

No network is trained on the target task.
For each target dataset (CIFAR-10, CIFAR-100, ImageNet-val, ImageNet-V2), the labeled sample is split 80/20 into reference and evaluation portions using a fixed split seed~$42$.
The $k$-NN decoder is built on the reference portion and evaluated on the held-out portion.
Because the representation is fixed independently of the target data, Q2 and Q3 vanish by construction and the certificate reduces to a single KL bound on the decoder error.

The pretrained backbones are listed in Table~1 of the main text.
For each backbone, the representation is extracted with a single deterministic forward pass using the preprocessing pipeline bundled with the official weights (see Appendix~\ref{appx:repr_details}).

\subsection{Table~2: Cross-fit certification of trained networks}
\label{app:table2}

This table evaluates the full three-term certificate described in Section~\ref{app:crossfit}, using the hyperparameters in Table~\ref{tab:hyperparams} and the architectures in Table~\ref{tab:architectures}.
Each configuration is run over multiple training seeds; the archived runs use seeds $\{0,1,2\}$ for MNIST and CIFAR-10 and seeds $\{0,1\}$ for CIFAR-100.
The table reports mean $\pm$ std over available seeds.

\subsection{Table~3: Comparison with prior bounds}
\label{app:table3}

This table is a positioning summary, not a unified rerun.
Baseline entries are copied from the corresponding publications; the ``Certified object'' column records what each method actually certifies (stochastic net, compressed net, surrogate, or deterministic net).

\paragraph{Source of ``Ours'' rows.}
All three ``Ours'' rows are cross-fit results, not transfer results:
\begin{itemize}[leftmargin=2em,itemsep=0pt]
  \item \textbf{MNIST:} ResNet-18 / ReLU, seeds $\{0,1,2\}$.
  \item \textbf{CIFAR-10:} ResNet-50 / ReLU, seeds $\{0,1,2\}$.
  \item \textbf{CIFAR-100:} ResNet-50 / ReLU, seeds $\{0,1\}$.
\end{itemize}

\subsection{Figure~1b: Sample-size sweep}
\label{app:fig1b}

A ReLU ResNet-18 is trained from scratch on CIFAR-10 subsets of size $n\in\{$1\,000, 5\,000, 10\,000, 20\,000, 30\,000, 40\,000, 50\,000$\}$.
Each subset is sampled uniformly at random (not stratified).
The training recipe is fixed across all subset sizes: SGD, lr~$0.1$, momentum~$0.9$, weight decay $5{\cdot}10^{-4}$, cosine decay, batch size~$128$, 200~epochs, standard crop-and-flip augmentation.
The full cross-fit certificate (Section~\ref{app:crossfit}) is evaluated at each $n$.
Seeds: $\{0,1,2\}$ (training), $42$ (fold partition).

\subsection{Figure~2: Width sweep and double descent}
\label{app:fig2}

Width-scaled ResNet-18 models are trained on CIFAR-10 at multipliers
$\{1, 2, 2.5, 3, 3.5, 4, 4.5, 5, 6, 8, 10, 12, 16, 20, 24, 32\}$.
(Exploratory runs at $48\times$ and $64\times$ were also trained but are omitted from the figure.)
The training recipe is held fixed: SGD, lr~$0.1$, momentum~$0.9$, cosine decay, batch size~$128$, 400~epochs.
Label noise is injected at levels 0\%, 10\%, and 20\% by replacing each selected label with a uniformly random class.
Seeds: $\{42,43,44\}$.

For each width, the plotted Q1 is the test-set decoder error at the $k$ minimizing Q1 over the grid; Q2 is the decoder--network disagreement at that same~$k$.
The figure therefore decomposes test-error into a monotone representation term (Q1) and a readout-agreement term (Q2) that peaks near the interpolation threshold.

\subsection{Figure~3: Causal geometry intervention}
\label{app:fig3}

\paragraph{Architecture and training.}
Six-layer FFNN, hidden width \textbf{1024}, trained on SVHN with AdamW (lr $10^{-4}$, weight decay $10^{-5}$), cosine decay, batch size 1024, 5\,000 epochs, seed~0.
Four activation functions are shown: ReLU, SiLU, GELU, tanh.

\paragraph{Regularizer schedule.}
The representation-repulsion regularizer $\mathcal{L}_{\mathrm{rep}}$ (Eq.~8 of the main text) uses a fixed strength $\lambda{=}2500$, identical across all four activations with no per-activation tuning.
The coefficient is held at zero through epoch~5, then increased linearly to its full value over the next 500 epochs (i.e.\ fully active after epoch~505).
The ramp allows the network to begin fitting the classification task before the repulsion term dominates.

\subsection{Figure~4: Weight-noise diagnostic}
\label{app:fig4}

\paragraph{Base model.}
This diagnostic uses a \emph{separate} CIFAR-10 ResNet-18 checkpoint, distinct from the cosine-decay checkpoints used in the certification tables.
It is trained with seed~42, SGD, lr~0.1, momentum~0.9, weight decay $5{\cdot}10^{-4}$, batch size~128, and a \textbf{multistep} schedule with drops by factor~$0.2$ at epochs~60, 120, and~160.
A different schedule was chosen for this diagnostic because the experiment predates the unified cosine-decay recipe adopted for the rest of the paper; the qualitative conclusion (representation degrades more slowly than the readout) is not schedule-dependent.

\paragraph{Noise protocol.}
Isotropic Gaussian noise $\mathcal{N}(0,\sigma^2)$ is added independently to every trainable parameter.
The noise levels are
$\sigma\in\{10^{-4},\,3{\cdot}10^{-4},\,10^{-3},\,0.002,\,0.003,\,0.005,\,0.008,\,0.01,\,0.015,\,0.02,\,0.03,\,0.05,\,0.1,\,0.2,\,0.3,\,0.5,\,1.0\}$.
For each $\sigma$ the noisy model is sampled \textbf{10 times}; both network accuracy and $k$-NN decoder accuracy ($k{=}5$, fixed 40\,000/10\,000 reference/query split of the CIFAR-10 training set) are averaged over these 10 draws.
The second panel fixes $\sigma{=}0.01$ and varies activation function and label-corruption level.

This figure is purely diagnostic and does not contribute to any formal certificate.

\subsection{Table~4: Desiderata comparison}
\label{app:table4}

Table~4 is qualitative.
It records structural properties of each certification approach (deterministic vs.\ stochastic, modified training, compression, surrogates, transfer applicability, architecture-agnostic proof, diagnostic interpretability).
No experimental detail is needed beyond what is stated in the table itself.

\subsection{Seeds and compute}
\label{app:seeds-compute}

Table~\ref{tab:seeds} collects all random seeds.

\begin{table}[h]
\centering\small
\caption{Random seeds used in each experiment.}
\label{tab:seeds}
\begin{tabular}{lcc}
\toprule
Experiment & Training seeds & Fold / split seed \\
\midrule
Table~1 (transfer)                              & --                 & 42 \\
Table~2 (MNIST, CIFAR-10)                       & $\{0,1,2\}$       & 42 \\
Table~2 (CIFAR-100)                             & $\{0,1\}$         & 42 \\
Table~3 (``Ours'')                              & same as Table~2    & 42 \\
Figure~1b (sample-size sweep)                   & $\{0,1,2\}$       & 42 \\
Figure~2 (width sweep)                          & $\{42,43,44\}$    & 42 \\
Figure~3 (causal intervention)                  & $\{0\}$           & -- \\
Figure~4 (weight noise)                         & $\{42\}$          & -- \\
\bottomrule
\end{tabular}
\end{table}

\paragraph{Compute.}
All runs are single-GPU, single-model jobs.
Archived wall-clock metadata covers the following:
Figure~1b: ${\approx}\,10.8$ GPU-hours (21~runs);
Table~2: ${\approx}\,102.7$ GPU-hours (64~runs);
Figure~2: ${\approx}\,72.9$ GPU-hours (162~runs);
total logged: ${\approx}\,186$ GPU-hours.
GPU model names were not recorded in the saved artifacts.
Comparable timing metadata were not logged for the transfer, regularizer, and weight-noise experiments.


\section{Representation extraction details}
\label{appx:repr_details}

All representations are extracted by a single deterministic forward
pass with no modification to the network.  No gradient is computed.
Evaluation inputs use deterministic preprocessing only (resize,
center crop, channel normalization); no stochastic augmentation is
applied at extraction time.

\paragraph{Feedforward networks (FFNN).}
Our FFNNs consist of $L$ fully connected hidden layers followed by a
linear classification head.  The representation $h_{\bm\theta}(x)$ is
the post-activation output of the final hidden layer, i.e.\ the
$D$-dimensional vector that is fed into the softmax classifier.  For
the experiments in \cref{sec:crossfit} we use $D = 1024$ with $L = 6$
hidden layers.  The activation function (ReLU or SiLU) is applied
before readout; this is the same activation used throughout the
network.

\paragraph{ResNets (ResNet-18, ResNet-50, WRN-50-2).}
The representation is the output of the global average pooling layer,
flattened to a vector of dimension $D$.  For ResNet-18, $D = 512$;
for ResNet-50 and WRN-50-2, $D = 2048$.  Concretely, we register a
forward hook on \texttt{model.avgpool} and flatten the resulting
tensor to shape $(B, D)$.  The final \texttt{fc} layer is not part of
the representation.

\paragraph{Vision Transformers (ViT-B/16, ViT-L/16).}
For torchvision ViTs, the representation is the class token
(\texttt{[CLS]}) extracted after the final encoder layer norm
(\texttt{model.encoder.ln}).  This yields $D = 768$ for ViT-B/16 and
$D = 1024$ for ViT-L/16.  We register a forward hook on the layer
norm and slice the first token: \texttt{output[:, 0, :]}.

\paragraph{DINOv2 (DINOv2-B/14, DINOv2-L/14, DINOv2-g/14).}
DINOv2 models are loaded via \texttt{torch.hub} from the
\texttt{facebookresearch/dinov2} repository.  The forward pass
returns the class-token embedding directly, so no hook is needed.
Dimensions are $D = 768$ (B/14), $D = 1024$ (L/14), and $D = 1536$
(g/14).  All DINOv2 backbones are used frozen with no fine-tuning.

\paragraph{Preprocessing.}
All models use ImageNet-normalized inputs
($\mu = (0.485, 0.456, 0.406)$,
$\sigma = (0.229, 0.224, 0.225)$).
For pretrained torchvision models, we use the preprocessing pipeline
bundled with the official weights
(\texttt{weights.transforms()}), which applies a deterministic resize
and center crop to $224 \times 224$.
For DINOv2 and randomly initialized models, we apply
\texttt{Resize(256)}, \texttt{CenterCrop(224)},
\texttt{ToTensor()}, and the same normalization.
For CIFAR experiments, models trained from scratch use
$32 \times 32$ inputs with per-dataset normalization; stochastic
augmentation (random crop, horizontal flip) is applied during
training only and never during feature extraction or bound
evaluation.

\paragraph{Numerical precision.}
All $\ell_2$ distances for $k$-NN are computed in \texttt{float64}.
Tie-breaking is deterministic: for the transfer experiments, a
lexicographic perturbation of $10^{-12} \times
\text{(reference index)}$ is added to squared distances, so that
among equidistant neighbors the one with the smallest index wins.
For the cross-fit experiments, ties are broken by stable sort order
over reference indices, which produces the same lowest-index-wins
rule.  Majority-vote ties are broken in favor of the lowest class
label (\texttt{argmax} over a vote-count tensor).

\paragraph{Computational cost.}
Feature extraction requires a single forward pass per input per
model.  The dominant cost is $k$-NN search, which is $O(n_{\mathrm{ref}}
\cdot n_{\mathrm{query}} \cdot D)$ per value of~$k$.  For the
largest transfer experiment (ImageNet, $n_{\mathrm{ref}} = 40{,}000$,
$n_{\mathrm{eval}} = 10{,}000$, $D = 1536$), this takes approximately
30 seconds on a single A100 GPU.  All $k$ values share the same
distance matrix, so the total cost scales linearly in $|\mathcal{K}|$
only through the top-$k$ selection step.



\end{document}